\documentclass[pdflatex,sn-basic]{sn-jnl}

\usepackage{optidef}
\usepackage{dsfont}
\usepackage{subcaption}

\jyear{2022}%

\theoremstyle{thmstyleone}%
\theoremstyle{thmstyletwo}%

\theoremstyle{thmstylethree}%

\begin{document}

\title[Gaussian Process regression over discrete probability measures: on the non-stationarity relation between Euclidean and Wasserstein Squared Exponential Kernels]{Gaussian Process regression over discrete probability measures: on the non-stationarity relation between Euclidean and Wasserstein Squared Exponential Kernels}


\author*[1]{\fnm{Antonio} \sur{Candelieri}}\email{antonio.candelieri@unimib.it}

\author[1,3]{\fnm{Andrea} \sur{Ponti}}\email{andrea.ponti@unimib.it}

\author[2]{\fnm{Francesco} \sur{Archetti}}\email{francesco.archetti@unimib.it}

\affil*[1]{\orgdiv{Department of Economics, Management and Statistics}, \orgname{University of Milano-Bicocca}, \orgaddress{\city{Milan}, \country{Italy}}}

\affil[2]{\orgdiv{Department of Computer Science, Systems and Communication}, \orgname{University of Milano-Bicocca}, \orgaddress{ \city{Milan}, \country{Italy}}}

\affil[3]{\orgname{OAKS srl}, \orgaddress{\city{Milan}, \state{Italy}, \country{Country}}}


\abstract{Gaussian Process regression is a kernel method successfully adopted in many real-life applications. Recently, there is a growing interest on extending this method to non-Euclidean input spaces, like the one considered in this paper, consisting of probability measures. Although a Positive Definite kernel can be defined by using a suitable distance -- the Wasserstein distance -- the common procedure for learning the Gaussian Process model can fail due to numerical issues, arising earlier and more frequently than in the case of an Euclidean input space and, as demonstrated in this paper, that cannot be avoided by adding artificial noise (\textit{nugget effect}) as usually done. This paper uncovers the main reason of these issues, that is a non-stationarity relationship between the Wasserstein-based squared exponential kernel and its Euclidean-based counterpart. As a relevant result, the Gaussian Process model is learned by assuming the input space as Euclidean and then an algebraic transformation, based on the uncovered relation, is used to transform it into a non-stationary and Wasserstein-based Gaussian Process model over probability measures. This algebraic transformation is simpler than \textit{log-exp} maps used in the case of data belonging to Riemannian manifolds and recently extended to consider the pseudo-Riemannian structure of an input space equipped with the Wasserstein distance.}




\maketitle

\section{Introduction}\label{sec1}

\subsection{Motivation}\label{sec1.1}
Gaussian Process (GP) regression \citep{williams2006gaussian,gramacy2020surrogates} is a well-known \textit{kernel method} \citep{scholkopf2018learning} widely adopted in Machine Learning (ML). One of its more successful application is Bayesian Optimization (BO) \citep{frazier2018bayesian,archetti2019bayesian,candelieri2021gentle}, but also optimal control and Reinforcement Learning \citep{deisenroth2013gaussian,sui2015safe,berkenkamp2016safe,jaquier2020bayesian}.

A GP regression model is typically learned (aka fitted) on a set of data-points laying into a vector space, equipped with the Euclidean distance.
However, there is recently an increasing interest on extending GP regression to non-Euclidean input spaces. The main motivation is that for many real-life applications, such as shape analysis,  diffusion tensor imaging, and robotic, the data often belongs to a non-Euclidean (aka \textit{non-flat}) \textit{manifold}, and ignoring that unfortunately leads to a GP regression model working on an incorrect geometric space. \citep{calandra2016manifold,Mallasto_2018_CVPR,jaquier2020bayesian}. Indeed, in the context of kernel methods, the so-called \textit{kernel function} encodes the -- typically non-linear -- similarity between two data-points, depending on their Euclidean distance. However, when the two elements to compare belong to a \textit{non-flat} manifold, the Euclidean distance is not the most appropriate any longer \citep{feragen2015geodesic,feragen2016open,Mallasto_2018_CVPR,jaquier2020bayesian}.\\

An interesting case, which is specifically considered in this paper, is when data belongs to a space whose elements are \textit{probability measures} -- i.e., probability distributions -- instead of data-points into a vector space. The \textbf{Wasserstein distance}, based on the \textbf{Optimal Transport (OT) theory} \citep{peyre2019computational,villani2021topics}, is a powerful tool to measure the distance between two distributions, with a wide range of applications \citep{simon2020barycenters,chen2022computing,ponti2021wasserstein,ponti2022unreasonable,candelieri2022use,candelieri2022distributional}. A relevant example is the design and development of effective and efficient Neural Architecture Search (NAS) approaches in the Deep Learning community \citep{kandasamy2018neural,nguyen2021optimal}. The aim of NAS is to efficiently search for neural architectures leading to accurate neural networks, but the conventional Euclidean distance usually fails in capturing differences between two architectures, which is at the core of the -- typically BO based -- search mechanism.\\

The critical issue in considering the Wasserstein distance in kernel methods, is that it is generally \textit{not-negative definite}, which may limit its ability to build Positive Definite (PD) kernels. As better detailed in the section devoted to related works (Sec.\ref{sec1.2}), many recent studies have proposed families of PD kernels based on the Wasserstein distance, under certain assumptions or based on some variants of the original distance.

Although a PD kernel can be defined by using the Wasserstein distance, we will empirically demonstrate that well-known numerical issues, occurring in learning a GP model, arise more frequently when data are probability measures.
Searching for a relationship between the Wasserstein-based
Squared Exponential (WSE) kernel and its Euclidean-based counterpart, we discovered a \textit{non-stationarity} relation linking the two kernels and this relation will be exploited to overcome the numerical issues. More precisely, a GP model with the SE kernel is learned as usual and then a simple algebraic transformation is used to obtain the \textit{equivalent WSE kernel}.
This algebraic transformation is less complicated than log-exp maps typically adopted in the case of data belonging to Riemannian manifolds/spaces, and which require significant modifications to be applied on data laying into an input space equipped with the Wasserstein distance (i.e., a pseudo-Riemannian manifold).

\subsection{Related works}\label{sec1.2}
In \citep{jaquier2020bayesian} a BO framework working on Riemannian manifolds has been recently proposed. The requirement emerges from a real-life application in robot-learning, with an objective function defined on a non-Euclidean input space. The Riemannian nature allows to use exponential and logarithmic maps (aka log-exp maps) for learning a suitable GP model of the objective function. Previously, in \citep{Mallasto_2018_CVPR} a \textit{wrapped} GP regression method was proposed to deal with data belonging to Riemannian manifolds, while the specific topic of positive definitiness (PDness) of the kernel has been investigated in \citep{feragen2015geodesic,feragen2016open}, with a focus on estimating the probability to obtain a PD \textit{geodesic} exponential kernel given a set of data laying on a non-Euclidean manifold. The term \textit{geodesic} means that the kernel between two data-points is computed according to their shortest path over the non-Euclidean manifold, instead of their Euclidean distance.

In \citep{bigot2017geodesic,cazelles2018geodesic} the Principal Component Analysis (PCA) is addressed in the case of an input space consisting of probability measures. Due to the nature of its elements, the input space must be equipped with an appropriate distance (i.e., the Wasserstein distance) which, however, makes it non-Euclidean and pseudo-Riemannian. The proposed approach uses the \textit{geodesics} (i.e., the shortest path between two distributions, according to the metrics) to take into account the non-linearity of the space. The approach is named Geodesic PCA (GPCA).
Similar topics are investigated in \citep{bachoc2017gaussian} and \citep{oh2019kernel,zhang2019optimal,de2020wasserstein} but, respectively, for GP regression on distribution inputs and, more generally, for Wasserstein-based kernel functions.\\

Finally, \textit{stationarity} is a typical assumption in GP regression, meaning that the same kernel function is used throughout the entire input space. In many real-world problems such an assumption could be not desirable because the modelled process might exhibit a different variability from one region to another of the input space. Although non-stationary kernels have been propoesed to overcome this limitation \citep{higdon1999non,schmidt2003bayesian,hebbal2021bayesian}, they result computationally impracticable. An efficient alternative, called Treed-GP, consists into partitioning the input space into subregions, and fitting separate stationary GP models within each subregion, leading
to a single non-stationary model \citep{kim2005analyzing,gramacy2008bayesian}. Although the first Treed-GP approaches date back to more than 10 years ago\citep{gramacy2007tgp,gramacy2011mesh}, a renewed interest on this topics has been recently emerging in different application domains \citep{civera2017detection,civera2020treed,candelieri2021treed}. In \citep{dolgov2018distance} a GP regression approach over distributions was proposed: the framework can also use non-stationary covariance functions.

\subsection{Contributions}\label{sec1.3}
\noindent
The main contributions of the paper are:
\begin{itemize}
    \item formalization and proof of a \textbf{non-stationarity} relation between the Wasserstein and the Euclidean Squared Exponential kernels. The relation is easier than the log-exp maps and it is directly learned from data;
    \item empirical evidence that learning a GP can fail easier and more frequently if the WSE kernel is used (even if it is PD in the case of univariate probability measures). It is known that numeric precision leads any PD kernel to behave as a PSD one, but the \textit{nugget effect} workaround, so effective for learning a GP with Euclidean-based kernels, does not work for the WSE kernel;
    \item a novel procedure for learning a WSE kernel-based GP on a dataset consisting of univariate discrete probability measures avoiding failure. The proposed procedure exploits the discovered \textbf{non-stationarity} relation: a GP model with SE kernel is learned as usual and then it is mapped into the equivalent non-stationary, WSE kernel-based, one;
    \item both theoretical proofs and empirical results on a set of test problems.
\end{itemize}

\subsection{Organization of the paper}\label{sec1.4}
The rest of the paper is organized as follows: Sec. \ref{sec2} summarizes the methodological background, relatively to GP regression (on Euclidean input space) and Wasserstein distance between probability measures. Sec. \ref{sec3} is the core of the paper, providing theoretical demonstrations of the \textit{equality} between Euclidean and Wassertein-based kernels, under certain assumptions. Sec. \ref{sec4} describes how the previous theoretical results are used to overcome computational issues in learning a GP model over probability measures. Sec. \ref{sec5} reports a set of examples to also provide empirical evidence of the benefits of the approach. Finally, we provide relevant conclusions, limitations, and perspectives of the research.

\section{Background}\label{sec2}

\subsection{Gaussian Process Regression}
\label{sec2.1}
A Gaussian Process (GP) can be though as an extension of the Gaussian distribution, working on functions instead of scalar values. Analogously to a Gaussian distribution, which is completely defined by its mean and the variance, a GP is completely defined by its mean and covariance \textit{functions}, respectively denoted with $\mu(\mathbf{x})$ and $k(\mathbf{x},\mathbf{x}')$.

From a ML perspective, GP regression belongs to the class of the kernel methods because most of the well-known kernel functions can be used as valid covariance functions. Different kernels lead to different structural assumptions (i.e., smoothness) of the resulting regression model, which is fitted (aka learned, trained, conditioned) on an available dataset $\mathcal{D}=\{\mathbf{X},\mathbf{y}\}$, with $\mathbf{X}\in \mathbb{R}^{N \times m}$ and $\mathbf{y} \in \mathbb{R}$. A generic row $\mathbf{x}_{i}$ of $\mathbf{X}$ is an $instance$, that is a data-point in a $m$-dimensional vector space, with associated $target$ value $y_i$. The most general case refers to possibly noisy targets, that is $y_i = f(\mathbf{x}_i)+\varepsilon_i$, with $\varepsilon_i\sim \mathcal{N}(0,\boldsymbol{\lambda}^2)\;,\forall i=1,...,N$.

Contrary to other \textit{deterministic} kernel methods, such as Support Vector Regression \citep{scholkopf2018learning}, a GP regression model is \textit{probabilistic} because it provides both a prediction and the associated uncertainty, for every possible input $\mathbf{x}$.
Prediction and uncertainty are given by the two following equations:
\begin{equation}
    \label{eq:gp}
    \begin{split}
    \mu(\mathbf{x}) & =  \mathbf{k}({\mathbf{x},\mathbf{X}})^\top \left[\mathbf{K}+\boldsymbol{\lambda}^2\mathbf{I}\right]^{-1} \mathbf{y} \\   
    \sigma^2(\mathbf{x}) & = k(\mathbf{x},\mathbf{x}) - \mathbf{k}(\mathbf{x},\mathbf{X})^\top \left[\mathbf{K}+\boldsymbol{\lambda}^2\mathbf{I}\right]^{-1} \mathbf{k}(\mathbf{x},\mathbf{X})
    \end{split}
\end{equation}

where $\mathbf{k}(\mathbf{x},\mathbf{X})$ is a $N$-dimensional (column) vector whose $i$-th component is $k(\mathbf{x},\mathbf{x}_i)$, $\mathbf{K}$ is a symmetric $N \times N$ matrix with entries $\mathbf{K}_{ij}=k(\mathbf{x}_i,\mathbf{x}_j)$, and the symbol $^\top$ denotes the transpose operator.   

Learning a GP model means that the two equations in (\ref{eq:gp}) must be conditioned to the available dataset $\mathcal{D}$. This is achieved by tuning the hyperparameters of the underlying kernel function.
The most common procedure is the maximization of the Marginal Likelihood Estimation (MLE), that is:
\begin{equation}
    \label{eq:mle}
    \underset{\boldsymbol{\theta}}{\arg \max} \left\{ -\frac{1}{2} \mathbf{y}^\top \left[\mathbf{K+\lambda^2\mathbf{I}}\right]^{-1} \mathbf{y} -\frac{1}{2} \log\left[\text{det}(\mathbf{K+\lambda^2\mathbf{I}})\right] - \frac{N}{2}\log 2\pi \right\}
\end{equation}

where $\boldsymbol{\theta}$ denotes the vector of the kernel's hyperparameters, which is involved in the computation of $\mathbf{K}$. An alternative is maximizing the Maximum-A-Posteriori (MAP) estimate \citep{williams2006gaussian,gramacy2020surrogates}.\\

As far as the choice of the kernel function is concerned, it is important to recall some relevant properties and definition:
\begin{itemize}
    \item a kernel is said to be \textit{stationary} if its value $k(\mathbf{x},\mathbf{x}')$ is a function of $\mathbf{x}-\mathbf{x}'$. Thus, it is invariant to translations;
    \item a kernel function is \textit{isoptropic} if its value $k(\mathbf{x},\mathbf{x}')$ is a function of only $\vert\mathbf{x}-\mathbf{x}'\vert$. Thus, it is invariant to all rigid motions;
    \item a kernel is said to be \textit{symmetric} if $k(\mathbf{x},\mathbf{x}')=k(\mathbf{x}',\mathbf{x})$. Clearly, only symmetric kernels can be used as valid covariance functions, because covariance is symmetric by definition;
    \item a kernel is said to be \textit{positive semidefinite} (PSD) if the associated kernel matrix, $\mathbf{K}$, is PSD for every $\mathcal{D}$ and every size $N \in \mathbb{N}$. The matrix $\mathbf{K}$ is PSD if and only if all of its eigenvalues are non-negative. This can also be written as $\mathbf{x}^\top \mathbf{K} \mathbf{x} \geq 0,\; \forall \mathbf{x}\in\mathbb{R}^m$. Instead, $\mathbf{K}$ is said \textit{positive definite} (PD) if $\mathbf{x}^\top \mathbf{K} \mathbf{x} > 0,\; \forall \mathbf{x}\in\mathbb{R}^m \setminus \{\mathbf{0}\}$ (i.e., all the eigenvalues are positive).
\end{itemize}

Although many PD kernels are available -- like those used in this paper -- the numerical precision leads to PSD kernel matrices \textit{de-facto}. Consider the following simple kernel:
\begin{equation*}
    k(\mathbf{x},\mathbf{\bar{x}}) = e ^ {-\|\mathbf{x}-\mathbf{\bar{x}}\|^2}
\end{equation*}

clearly it is PD because $k(\mathbf{x},\mathbf{\bar{x}})=1 \iff \|\mathbf{x}-\mathbf{\bar{x}}\|=0$ and, consequently, every generated kernel matrix is PD, with entries $K_{ii}=1$ and $0<K_{ij}<1$. It is also important to remark that scaling every dimension through a vector $\boldsymbol{\theta} \in \mathbb{R}_+^m$ does not affect this result.

However, due to the limited numerical precision of the computers,  we have to consider that $\|\mathbf{x}-\mathbf{\bar{x}}\|\leq \epsilon$, with $\epsilon>0$, practically means $\|\mathbf{x}-\mathbf{\bar{x}}\| = 0$. Thus, if the two input are closer than the numerical precision then $k(\mathbf{x},\mathbf{\bar{x}})=1$, leading to entries $K_{ij}=1$ and, consequently, to a PSD -- not PD -- kernel matrix.

This numerical issue is well known in GP regression, especially arsing in the case of a noise-free $f(\mathbf{x})$. Indeed, a noise-free setting implies $\lambda^2=0$ in equation (\ref{eq:mle}), while a limited numerical precision, $\epsilon$, leads to a PSD $\mathbf{K}$ which cannot be inverted because $\det(\mathbf{K})=0$ (aka, \textit{ill-conditioning} of $\mathbf{K}$). Thus, the first two terms of (\ref{eq:mle}) cannot be computed.
This issue is usually solved by adding a \textit{nugget effect}, that is an \textit{artificial} noise -- even if $f(\mathbf{\mathbf{x}})$ is known being noise-free. Basically, instead of estimating $\lambda^2$ from noisy observations, it is artificially added to avoid the ill-conditioning of $\mathbf{K}$. This is the most widely adopted workaround in GP-based BO softwares.\\

It is important to remark that all the reported properties and considerations depend on how the distance $\mathbf{x}-\mathbf{x}'$ is computed. Most of the research focused on vector spaces and, consequently, on Euclidean (as in the example, that is  $\mathbf{x}-\mathbf{x}' = \|\mathbf{x}-\mathbf{x}'\|$), Manhattan, Mahalanobis, etc.
However, when the data are not data-points in a vector space, but probability distributions, a more suitable distance must be considered.

\subsection{Wassertein distance and Optimal Transport theory}

The Wasserstein distance is used to compare probability measures \citep{peyre2019computational,villani2021topics}.
From the Optimal Transport (OT) theory, the general formula is:
\begin{equation}
    \label{eq:Wp}
    \mathcal{W}_p(\boldsymbol{\alpha},\boldsymbol{\alpha}') = \mathcal{L}_{d^p}(\boldsymbol{\alpha},\boldsymbol{\alpha}')^{1/p}    
\end{equation}

where $d(z_i,z_j')$ is a distance measure between elements of the so-called \textit{supports}, $\mathbf{z}$ and $\mathbf{z}'$, of the two distributions $\boldsymbol{\alpha}$ and $\boldsymbol{\alpha}'$. The distance $d(z_i,z_j')$ is named \textit{ground metric}, on which it is based the computation of the cost incurred to move a quantity of \textit{mass} from the location $z_i$ of $\boldsymbol{\alpha}$ to the location $z_j'$ of $\boldsymbol{\alpha'}$, with the aim to match $\boldsymbol{\alpha}$ with $\boldsymbol{\alpha}'$.
Finally, $\mathcal{L}_{d^p}$ is the overall cost resulting from all the probability mass movements. Thus, it is clear why it is an optimal transport problem: the aim is to find the mass movements minimizing $\mathcal{L}_{d^p}$, that is identifying the \textit{optimal transport plan}.

When $d(z_i,z_j')$ is a distance\footnote{The distance between an object and itself is always zero; the distance between distinct objects is always positive;
it is symmetric; and it satisfies the triangle inequality}, then, also the $\mathcal{W}_p(\boldsymbol{\alpha},\boldsymbol{\alpha}')$ is a distance, for some $p\geq 1$. Moreover, compared to \textit{divergence measures}, such as Kullback-Leibler and Jensen-Shannon, Wasserstein is also more general, allowing to compare two discrete, two continuous, or one continuous and one discrete distributions, where the type of a distribution is defined according to the nature of its support (i.e., $\mathbf{z} \in \{v_1, ...., v_h\}$ for a discrete distribution, $\mathbf{z} \in \mathbb{R}^n$, for a continuous one). Along with the type of support, also its dimensionality is important, leading to different mathematical properties of the Wasserstein distance for univariate and multivariate probability distributions. Properties also depend on the parameter $p$, for which the most convenient choice is $p \in [1,2]$.\\

For the scope of this paper, we focus on \textit{univariate discrete probability measures} and the associated Wasserstein distance theory's results. For a more extensive methodological background about the Wasserstein distance on more general cases (e.g., different values of $p$, multivariate and/or continuous probability measures) the reader can refer, for instance, to \citep{peyre2019computational,villani2021topics}.

A univariate discrete probability measure $\boldsymbol{\alpha}$ is represented through two $m$-dimensional vectors:
\begin{itemize}
    \item the \textbf{support} $\mathbf{z}=(z_1,...,z_m)$, with $z_i \in \mathbb{R}, i=1,...,m$;
    \item the \textbf{weights} $\mathbf{a}=(a_1,...a_m)$, with $\mathbf{a} \in \Delta_{m-1}$
\end{itemize}

\noindent
with $\Delta_{m-1}$ the so-called \textbf{probability simplex}, that is:
\begin{equation*}
    \Delta_{m-1}=\left\{\mathbf{a} \in \mathbb{R}^m:\sum_{i=1}^m a_i=1\right\}  
\end{equation*}

In other words, $\boldsymbol{\alpha}=\sum_{i=1}^m a_i \delta_{z_i}$, with $\delta_z$ the Dirac's delta function.
Roughly speaking, $\forall\; i=1,...,m$, an amount of \textit{mass} $a_i$ is located at $z_i$, and the overall mass of $\boldsymbol{\alpha}$ is $1$.

As previously mentioned, $\mathcal{L}_{d^p}(\boldsymbol{\alpha},\boldsymbol{\alpha}')$, in (\ref{eq:Wp}), is computed by solving the following optimal transport problem (due to Kantorovich, in 1942), in the case of discrete probability measures:
\begin{mini}|s|
{\mathbf{T} \in {\mathcal{T}}} {\left [\sum_{ij}\left(\mathbf{C}\odot\mathbf{T}\right)_{ij}\right]^{1/p}}
{}{}
\addConstraint{\sum_{i=1}^d \mathbf{T}_{ij} = a'_j \quad \forall j}
\addConstraint{\sum_{j=1}^d \mathbf{T}_{ij} = a'_i  \quad \forall i}
\label{eq:P1}
\end{mini}

where $\mathbf{T} \in \mathbb{R}^{\vert \mathbf{z}\vert \times \vert \mathbf{z}' \vert}$ is a transport map/plan, such that $\mathbf{T}_{ij}$ is the amount of mass moved from $z_i$ to $z_j'$ to match $\boldsymbol{\alpha}$ with $\boldsymbol{\alpha}'$, $\mathcal{T} $ is the set of all the possible transport maps/plans, $\odot$ denotes the Hadamard product (i.e., element-wise product, that is $\mathbf{C}\odot\mathbf{T}=\mathbf{C}_{ij}\mathbf{T}_{ij}$), and $\mathbf{C}_{ij}=d^p(z_i,z'_j)$.\\

\noindent
In this paper we will focus on the following two settings:
\begin{itemize}
    \item $d^p(z_i,z_j')$ is the \textit{binary ground metric}, that is:
    \begin{equation}
        \label{eq:binary_ground_metric}
        d^p(z_i,z_j') = 
        \begin{cases}
            1, & \text{if}\ z_i=z_j' \\
            0, & \text{otherwise}
    \end{cases}
    \end{equation} 
    
    \item $d^p(z_i,z_j')$ can be any distance, specifically with $p=2$.
\end{itemize}

It is easy to demonstrate that, when the ground metric is binary, as defined in (\ref{eq:binary_ground_metric}), and $p=1$, then, the Wassertein distance between two discrete probability measures having the same identical support, $\mathbf{z}$, is equal to half of the Total Variation between their weight vectors \citep{peyre2019computational}:
\begin{equation}
    \label{eq:tv}
    \mathcal{W}_1(\boldsymbol{\alpha},\boldsymbol{\alpha}') =\frac{1}{2}\sum_{i=1}^d \vert a_i - a_i' \vert = \frac{TV(\mathbf{a},\mathbf{a}')}{2}
\end{equation}
This remark is important and it will be used later in the paper.

\subsection{Wasserstein versus Euclidean distance}\label{sec2.3}

Here we report simple examples showing differences between the Euclidean and the Wasserstein distance (\ref{eq:Wp}). For the sake of simplicity, we consider univariate discrete distributions having identical supports, with $m=2$.

\begin{figure}[h]
\begin{subfigure}{.5\textwidth}
    \centering
    \includegraphics[scale=0.27]{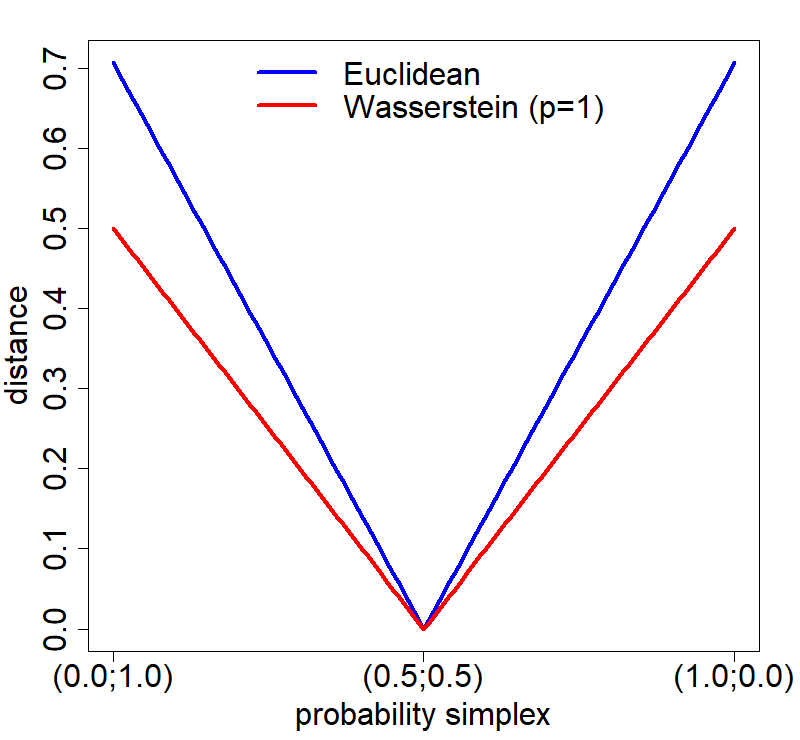}
\end{subfigure}
\begin{subfigure}{.5\textwidth}
    \centering
    \includegraphics[scale=0.27]{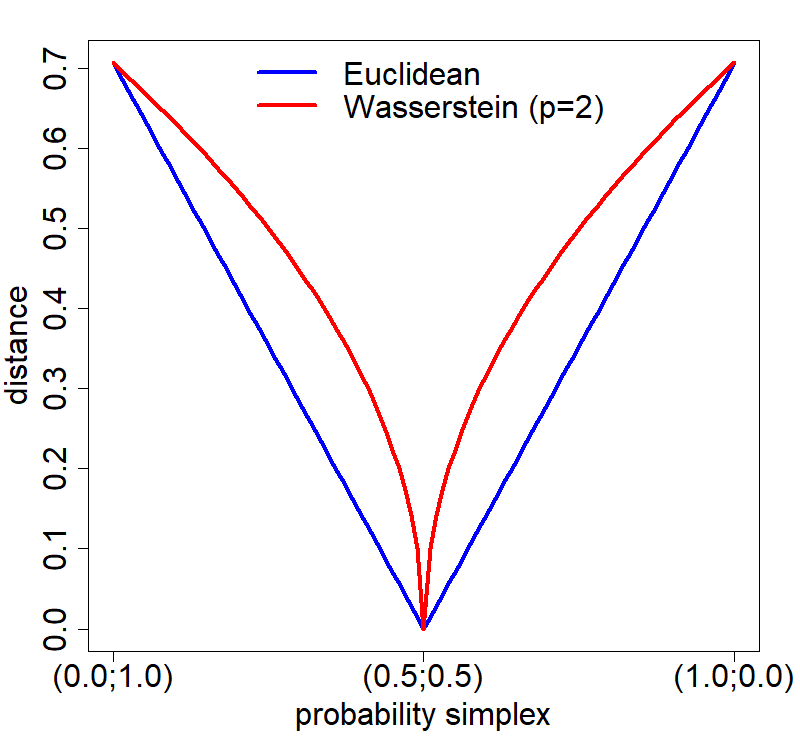}
\end{subfigure}
\caption{Univariate probability simplex: Euclidean distance versus Wasserstein distance with \textbf{binary ground metric} and $p=1$ (on the left) and $p=2$ (on the  right). Distances are from the reference probability measure having weight vector $\mathbf{a}_0=(0.5,0.5)$.}
\label{fig:f1}
\end{figure}

As a first example, we consider the distance between any other distribution of the unidimensional probability simplex and the \textit{reference distribution}, $\boldsymbol{\alpha}_0$, having weight vector $\mathbf{a}_0=(0.5; 0.5)$. The ground metric is the binary one. Figure \ref{fig:f1} shows the difference between the Euclidean and the Wasserstein distance in the case that $p=1$ (on the left hand side) and $p=2$ (on the right hand side).

It is important to notice how $p=2$ implies that Wasserstein is a non-linear distance over the probability simplex (unlike the Euclidean). On the other hand, for $p=1$ also the Wasserstein distance is linear and there exists a scalar relation between the two distances. Furthermore, while $\mathcal{W}_2$ is always greater or equal to the Euclidean distance, the opposite behaviour is observed for $\mathcal{W}_1$.
\vskip -0.25cm

\begin{figure}[h]
\begin{subfigure}{.5\textwidth}
    \centering
    \includegraphics[scale=0.27]{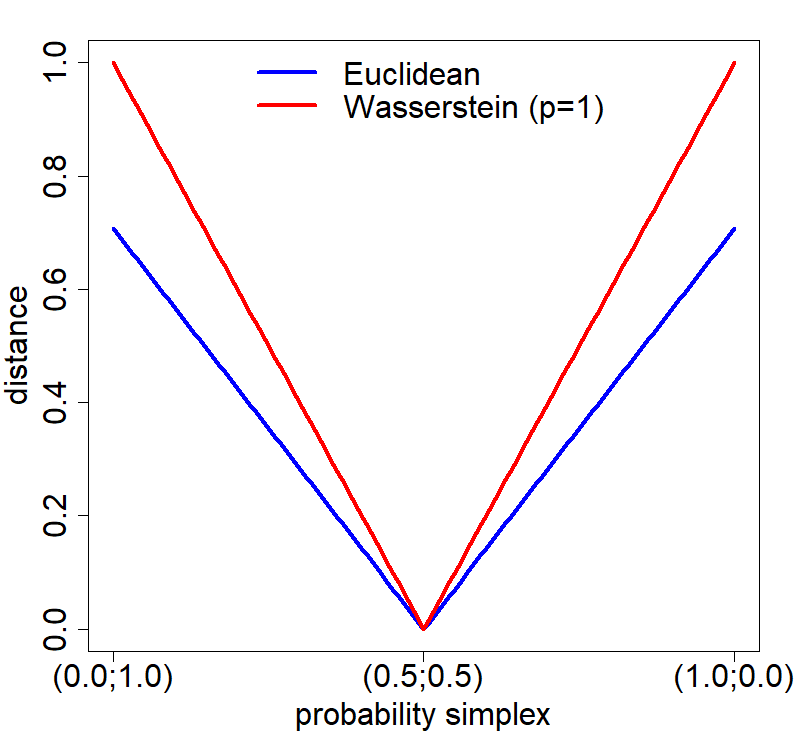}
\end{subfigure}
\begin{subfigure}{.5\textwidth}
    \centering
    \includegraphics[scale=0.27]{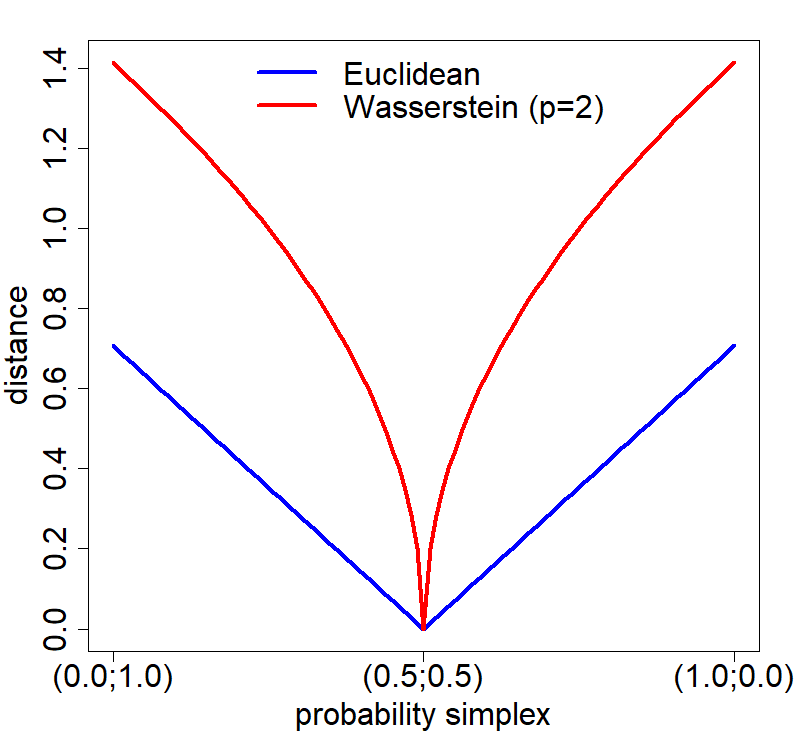}
\end{subfigure}
\caption{Univariate probability simplex: Euclidean distance versus Wasserstein distance with \textbf{Euclidean ground metric} and $p=1$ (on the left) and $p=2$ (on the  right). Distances are from the reference probability measure having weight vector $\mathbf{a}_0=(0.5,0.5)$.}
\label{fig:f2}
\end{figure}

As second example we simply modify the ground metric, now defined as the Euclidean distance between the support locations. Specifically, we decided to set $\mathbf{C}_{12} = \mathbf{C}_{21} = 2$ and, obviously, $\mathbf{C}_{11} = \mathbf{C}_{22} = 0$. The new situation is depicted in Figure \ref{fig:f2}. The new ground metric does not modify the type of (non-linear/linear) relation between the two distances, but an amplification effect is observed on Wasserstein, which is now greater than the Euclidean distance also in the case $p=1$. This amplification is due to the ground metric, specifically to the cost $\mathbf{C}_{ij}$; an opposite behavior (i.e., a decrease in the Wasserstein values) would be observed for $\mathbf{C}_{ij} \in [0, 1]$.

\newpage

\begin{figure}[h]
\begin{subfigure}{.5\textwidth}
    \centering
    \includegraphics[scale=0.27]{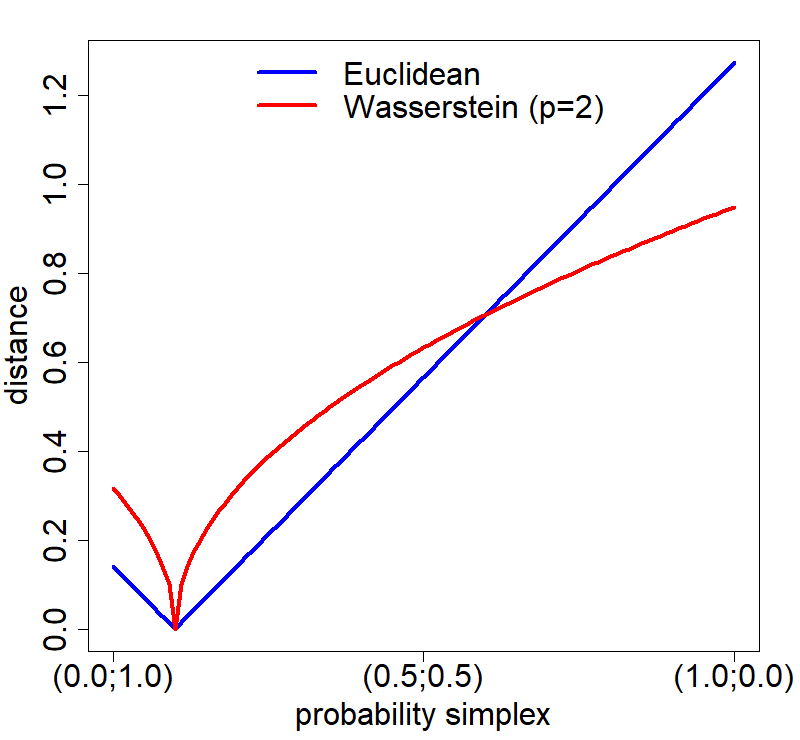}
\end{subfigure}
\begin{subfigure}{.5\textwidth}
    \centering
    \includegraphics[scale=0.27]{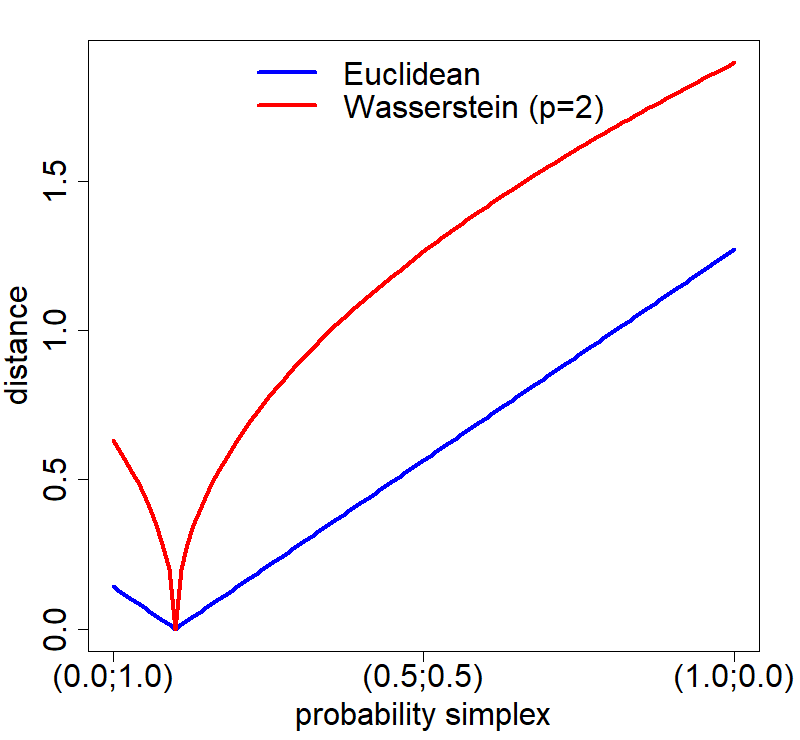}
\end{subfigure}
\caption{Univariate probability simplex: Euclidean distance versus Wasserstein distance with $p=2$ and with binary ground metric (on the left) and Euclidean ground metric (on the right). Distances are from the reference probability measure having weight vector $\mathbf{a}_0=(0.1,0.9)$.}
\label{fig:f3}
\end{figure}

Finally, the role of the reference distribution is considered, for the case $p=2$ only (for $p=1$ is trivial). Indeed, the weight vector of the new reference distribution, $\boldsymbol{\alpha}_0$, is now $\mathbf{a}_0=(0.1; 0.9)$ and Figure \ref{fig:f3} shows the difference between $\mathcal{W}_2$ and the Euclidean distance with respect to using (on the left hand side) the binary ground metric and (on the right hand side) the Euclidean-based ground metric previously defined.
\vskip -0.25cm

To better understand how equipping the input space with a non-Euclidean distance can drastically affect the structural properties of a function, we report here a simple example. Consider the following stationary, smooth, continuous 1-dimensional function:
\begin{equation}
    \label{eq:test1}
    f(x):[0,1] \rightarrow \mathbb{R}, 
    f(x) = sin(4.8x+2.7) + sin\left( \frac{10(4.8x+2.7)}{3}\right)
\end{equation}

\begin{figure}[h]
    \centering
    \includegraphics[scale=0.27]{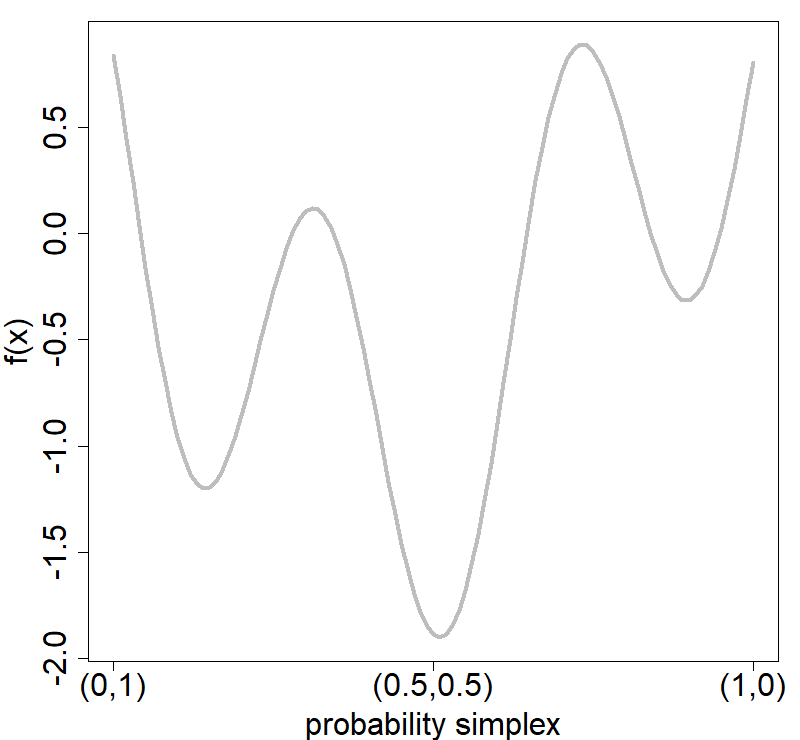}
    \caption{An example of a function defined over the probability simplex equipped with the Euclidean distance.}
    \label{fig:f4}
\end{figure}

It can be redefined over the 1-dimensional probability simplex of discrete probability measures ($m=2$), for instance by simply setting $f(x)=f(a_1)$, with $\mathbf{a}=(a_1,a_2)$, as depicted in Figure \ref{fig:f4}.
\vskip -0.25cm

\begin{figure}[h]
\begin{subfigure}{.5\textwidth}
    \centering
    \includegraphics[scale=0.27]{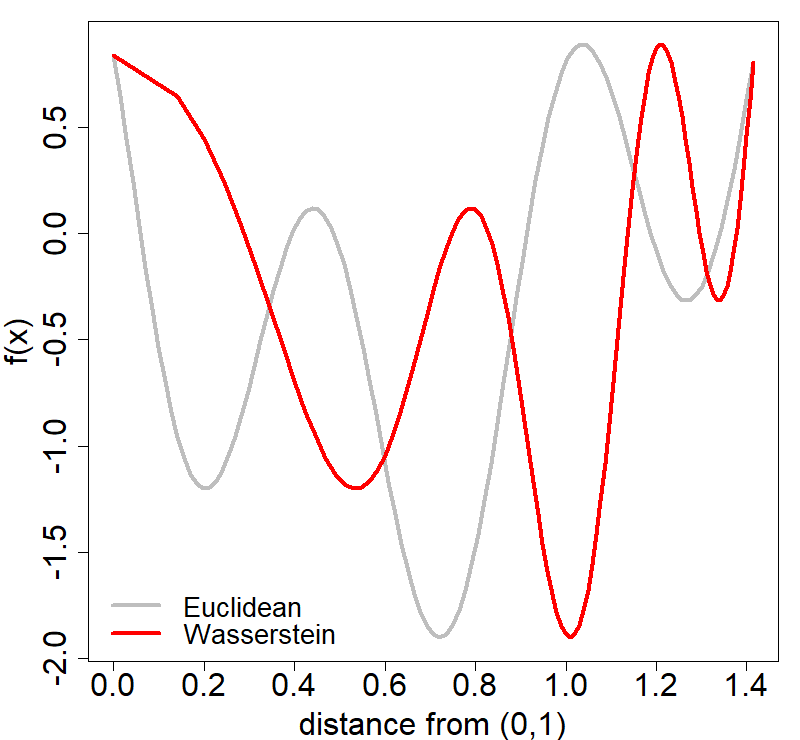}
\end{subfigure}
\begin{subfigure}{.5\textwidth}
    \centering
    \includegraphics[scale=0.27]{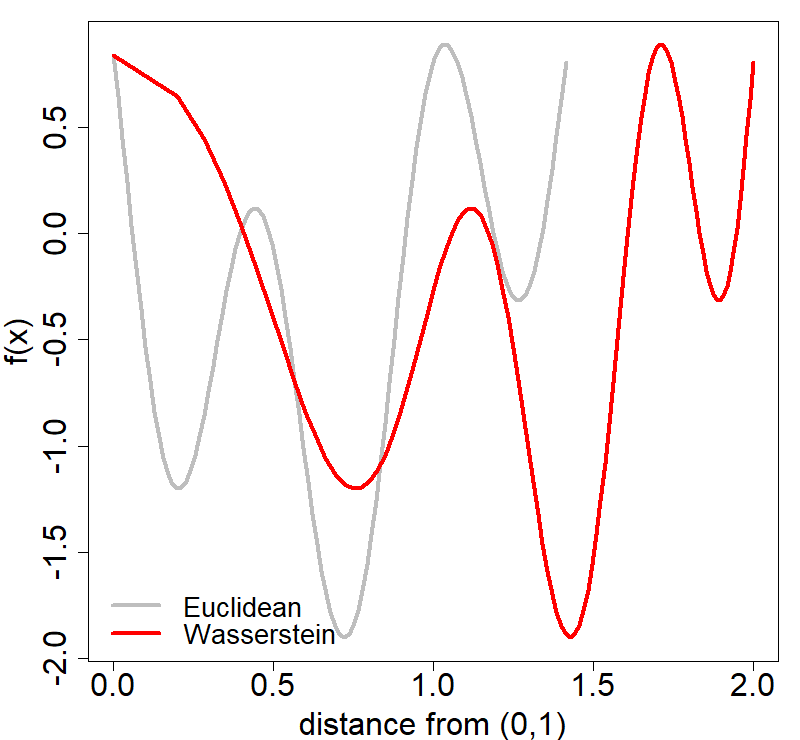}
\end{subfigure}
\caption{Comparison between the shapes of the function (\ref{eq:test1}) when the univariate probability simplex is equipped with the Euclidean (grey) and Wasserstein (red) distance. For the Wasserstein distance, $p=2$ and the ground metric is Euclidean: on the left hand side $\mathbf{C}_{ij}=2$, on the right hand side is $\mathbf{C}_{ij}=4$.}
\label{fig:f5}
\end{figure}

The shape of the function becomes completely different if the probability simplex is equipped with the Euclidean or the Wasserstein distance. Consider the distance from the distribution at the left-hand side of the simplex, that is $(0,1)$, then Figure \ref{fig:f5} makes evident the differences between the two cases (i.e., for the Wasserstein distance we set $\mathbf{C}_{ij}=2$, on the left chart, and $\mathbf{C}_{ij}=4$, on the right chart). Two important points arise:
\begin{itemize}
    \item smoothness becomes not constant over the input space when equipped with the Wasserstein distance. Indeed, the function becomes \textbf{non-stationary}.
    \item the maximum of the Wasserstein distance from the reference distribution -- $(0,1)$, in our case -- can be equal, larger, as well as smaller than the maximum Euclidean distance, depending on the ground metric, respectively $\mathbf{C}_{ij}=2$, $\mathbf{C}_{ij}>2$, and $\mathbf{C}_{ij}<2$, in our example (for the sake of visualization, only the equal and larger cases are depicted).
\end{itemize}
\vskip -0.25cm

\newpage

We conclude this section with an important property of the Wasserstein distance.
Although $\mathcal{W}_p$ with $p>1$ is non-linear over the probability simplex, in the case of univariate probability distributions (also with different supports) it has a closed form, simply considering the \textit{pseudoinverse of the cumulative distribution function} \citep{peyre2019computational}. This leads to an important result: \textbf{the Wasserstein distance for univariate probability measures is a Hilbertian metric} (it is not Hilbertian for multi-dimensional supports, that is multi-variate probability measures) \citep{bachoc2017gaussian,peyre2019computational}.
With respect to this result, it is important to remark that Hilbertian distances can be cast as a PD (radial basis function) kernel:
\begin{equation}
    k(\cdot,\cdot) = e^{-\frac{\left(d\left(\cdot,\cdot\right)\right)^h}{t}}
\end{equation}
with $0\leq h \leq 2$ and $t>0$.

\section{On the equivalence between Euclidean and Wasserstein Squared Exponential kernels}\label{sec3}

In this section, we detail the main theoretical result from our study. First we introduce (\textit{a}) the (Euclidean) Squared Exponential (SE) kernel, working on weight vectors -- so, computing the similarity between two distributions depending on the Euclidean distance between their weight vectors -- and (\textit{b}) the Wassertein Squared Exponential (WSE) kernel -- computing the similarity between two distributions depending on their Wasserstein distance.
Then, we provide two \textit{equality relations} between the two kernels, under two different conditions. These results will usher the most relevant outcome about non-stationarity relationship between the GPs model using the two different kernels, (presented later in Sec. \ref{sec4}).

\subsection{Kernels definition}\label{sec3.1}

The SE kernel is one of the mostly well-known kernel functions, widely adopted in kernel methods. In our case, it works on the weight vectors of two probability distributions. This means that their similarity, encoded by the kernel, is computed depending on the Euclidean distance between their associated weight vectors, as follows:
\begin{equation}
    k_{SE}(\mathbf{a},\mathbf{a}') = e^{-\frac{\|\mathbf{a}-\mathbf{a}'\|^2_2}{2\ell^2}}
    \label{eq:kse}
\end{equation}

with $\ell$ the so-called \textit{length-scale}. When $\ell \in \mathbb{R}_+$ the kernel is said to be \textit{isotropic}, otherwise a different length-scale can be adopted for each one of the vector space's dimension, leading to $\boldsymbol{\ell} \in \mathbb{R}_+^m$ (i.e., anisotropic SE kernel).\\

Using the previous result about ``Hilbertianity'' of the Wasserstein distance (Sec. \ref{sec2.3}), we define the Wasserstein Squared Exponential (WSE) kernel by replacing the Euclidean distance between two weight vectors with the Wasserstein distance -- with $p=2$ -- between the two probability distributions:
\begin{equation}
    k_{\mathcal{W}SE}(\boldsymbol{\alpha},\boldsymbol{\alpha}') = e^{-\frac{\mathcal{W}_2^2(\boldsymbol{\alpha},\boldsymbol{\alpha}')}{2\rho^2}}
    \label{eq:kwse}
\end{equation}
with $\rho$ the length-scale of the WSE kernel: a different symbol is used to differentiate it from the previous $\ell$. Contrary to the SE kernel, the WSE kernel should be -- and usually it is -- considered isotropic, that is $\rho\in\mathbb{R}_+$. Indeed, every $\ell_i$ allows to rescale the distance along every dimensions of the input space, but it is difficult to find an analogy for $\rho_i$ since is is not associated to any dimension of the input space.\\

Usually, a further kernel's hyperparameter is also used, namely $\sigma_f^2$, that is a multiplier regulating the variation in amplitude of the kernel. Instead of varying in $[0,1]$, the values of kernel ranges in $[0,\sigma_f^2]$. From now on we omit $\sigma_f^2$ (i.e., we assume $\sigma_f^2=1)$ for both the kernels, without any loss of generality.

\subsection{Equivalence via ground metric modification}\label{sec3.2}

Our study starts by demonstrating under which conditions, and also how, it is possible to modify the ground metric to make the two kernels, (\ref{eq:kse}) and (\ref{eq:kwse}), equal one to each other.\\

\textbf{Theorem 1.}
\textit{In the case of univariate discrete probability measures, with the same support, and under a binary ground metric, it is possible to compute a modified ground metric, $\widetilde{d}^2(z_i,z_j')$, making $\mathcal{W}_2^2(\boldsymbol{\alpha},\boldsymbol{\alpha}')=\|\mathbf{a}-\mathbf{a}'\|^2_2$}.\\

\textbf{Proof.} The binary ground metric defined in (\ref{eq:binary_ground_metric}) leads to a binary cost matrix $\mathbf{C}$ such that $\mathbf{C}_{ii}=0$ and $\mathbf{C}_{ij}=1$. Solving the OT problem results into the optimal transport plan, $\mathbf{T}^*$, with the associated overall transportation cost equal to $\mathcal{W}_p(\boldsymbol{\alpha},\boldsymbol{\alpha}')=\left[\mathbf{C}\odot\mathbf{T}^*\right]^{1/p}=\left[\sum_{ij}\mathbf{C}_{ij}\mathbf{T}^*_{ij}\right]^{1/p}$.

Note that the values of $\mathbf{C}_{ii}$ and $\mathbf{C}_{ij}$ are not affected by $p$, because $d^p(z_i,z_j')$ is the binary ground metric.\\

Now, we introduce a modified cost matrix, $\mathbf{\widetilde{C}}$ such that $\widetilde{\mathbf{C}}_{ij}=\mathbf{C}_{ij}\xi_i$. In a more compact form, we can write $\mathbf{\widetilde{C}}=\mathbf{C}\odot\boldsymbol{\xi} \mathbf{1}_m^\top$.

It is trivial to demonstrate that using $\mathbf{\widetilde{C}}$ instead of $\mathbf{C}$ does not change the OT plan $\mathbf{T}^*$, but just its final cost:
\begin{equation}
\begin{split}
    \left[\mathbf{\widetilde{C}}\odot\mathbf{T}^*\right]^{1/p} & = \left[ \sum_{ij} \widetilde{C}_{ij} T^*_{ij} \right]^{1/p} =\\
    & = \left[ \sum_i \left( \sum_j \mathbf{C}_{ij} \mathbf{T}^*_{ij} \right) \xi_i \right]^{1/p}
\end{split}
\end{equation}

\noindent
Now we have to impose $\mathcal{W}_2^2(\boldsymbol{\alpha},\boldsymbol{\alpha}') = \|\mathbf{a}-\mathbf{a}'\|^2_2$; thus, $p=2$ leading to:
\begin{equation}
    \label{eq:equality}
    \sum_i \left( \sum_j \mathbf{C}_{ij} \mathbf{T}^*_{ij} \right) \xi_i = \sum_{i=1}^m (a_i - a_i')^2
\end{equation}

where the left hand side term is $\mathcal{W}_2^2(\boldsymbol{\alpha},\boldsymbol{\alpha}')$ and the right hand side term is $\|\mathbf{a}-\mathbf{a}'\|^2_2$.
It is important to notice that:
\begin{equation*}
    a_i-a_i'=0 \implies \sum_j \mathbf{C}_{ij}\mathbf{T}^*_{ij}=0
\end{equation*}
but 

\begin{equation*}
    \mathbf{C}_{ij}\mathbf{T}^*_{ij}=0 \;\not\!\!\!\implies a_i-a_i'=0
\end{equation*}
thus, we cannot simply impose: 
\begin{equation*}
    \left( \sum_j \mathbf{C}_{ij} \mathbf{T}^*_{ij} \right) \xi_i = (a_i - a_i')^2, \quad \forall i
\end{equation*}

because this does not satisfy (\ref{eq:equality}).
Therefore, we have to introduce the following set $\mathcal{I}$:
\begin{equation}
    \label{eq:setI}
    \mathcal{I} = \left\{ i: \sum_j \mathbf{C}_{ij}\mathbf{T}_{ij}=0  \right\}
\end{equation}
from which we compute the quantity
\begin{equation*}
    q = \sum_{i \in \mathcal{I}} (a_i-a_i')^2
\end{equation*}
Then, we can obtain $\boldsymbol{\xi} \in \mathbb{R}_+^m$ as follows:
\begin{equation}
    \label{eq:xi}
    \xi_i = 
    \begin{cases*}
        \frac{(a_i-a_i')^2+q/(m-\vert\mathcal{I}\vert)}{\sum_j \mathbf{C}_{ij}\mathbf{T}^*_{ij}} & if $i \notin \mathcal{I}$\\
        \quad\quad \star & otherwise
    \end{cases*}
\end{equation}
where $\star$ denotes whatever value, as it does not contribute to the overall transportation cost because the associated $\sum_j \mathbf{C}_{ij}\mathbf{T}^*_{ij}=0$ by definition.
Finally, the modified ground metric $\widetilde{d}^2(z_i,z_j')$ (i.e., we have set $p=2$), is given by:
\begin{equation}
    \label{eq:modified_d}
    \widetilde{d}^2(z_i,z_j') = 
    \begin{cases*}
        0 & if $i=j$ (and $\mathbf{z}=\mathbf{z}'$)\\
        \xi_i & as in (\ref{eq:xi}) otherwise
    \end{cases*}
\end{equation}

It is important to remark that computing the modified ground metric $\widetilde{d}^2(z_i,z_j')$ requires to preliminary solve the OT problem, with $d(z_i,z_j')$ the binary ground metric (\ref{eq:binary_ground_metric}), in order to obtain the optimal transport plan $\mathbf{T}^*$.
Moreover, $\mathbf{a}$ and $\mathbf{a}'$ are needed for computing $\widetilde{d}^2(z_i,z_j')$ -- specifically $\xi_i$. This means that $\xi_i$ is, actually, a function of $\mathbf{a}$ and $\mathbf{a}'$.\\

Finally, it is important to clarify that, according to (\ref{eq:xi}), the quantity $q$ is equally distributed over $m-\vert\mathcal{I}\vert$ different $\xi_i$, with $i\notin\mathcal{I}$. This is the easiest choice, 
but it is not unique.\\

\textbf{Corollary 1.1.} \textit{Following from Theorem 1, and under the same assumptions, the SE and the WSE kernels, both isotropic and having the same length-scale, are equivalent}.\\

\noindent
\textbf{Proof.} From Th.1 we know that 
\begin{equation*}
    \exists\; \widetilde{d}^2(z_i,z_j): \mathcal{W}_2^2(\boldsymbol{\alpha},\boldsymbol{\alpha}')=\|\mathbf{a}-\mathbf{a}'\|^2_2
\end{equation*}
\noindent
with $\widetilde{d}^2(z_i,z_j')$ as defined in (\ref{eq:modified_d}). Moreover, assuming $\ell = \rho \in \mathbb{R}_+$, it follows: 
\begin{equation*}
    k_{SE}(\mathbf{a},\mathbf{a}') = e^ {- \frac{\|\mathbf{a}-\mathbf{a}'\|^2_2}{2 \ell ^2}} = e^ {- \frac{\mathcal{W}^2_2(\boldsymbol{\alpha},\boldsymbol{\alpha}')}{2 \rho^2}} = k_{\mathcal{W}SE}(\boldsymbol{\alpha},\boldsymbol{\alpha}') 
\end{equation*}
QED.\\

The previous Corollary is only valid if the two kernels are isotropic. Anyway, it is trivial to demonstrate that this result can be extended to anisotropic kernels by simply setting $\ell_i=\rho_i, \forall i=1,...,m$.
However, as already mentioned, the role of $\rho_i$ is less intuitive than $\ell_i$'s one. Indeed, $\ell_i$, in the SE kernel, rescales, for each dimension, the distance between $\mathbf{a}$ and $\mathbf{a}'$ along that dimension. To understand the role of $\rho_i$, we have to consider that
\begin{equation}
    \label{eq:W_rho}
    \frac{\mathcal{W}_2^2(\boldsymbol{\alpha},\boldsymbol{\alpha}')}{\boldsymbol{\rho^2}} =
    \frac{\sum_{ij}\mathbf{C}_{ij}\mathbf{T}^*_{ij}}{\boldsymbol{\rho}^2}=\sum_i\left(\sum_j \frac{\mathbf{C}_{ij}\mathbf{T}^*_{ij}}{\rho_i^2} \right)
\end{equation}

meaning that the cost $\mathbf{C}_{ij}$ for moving a quantity of mass from the (source) location $z_i$ of $\boldsymbol{\alpha}$ to \textbf{any} (sink) location $z_j'$ of $\boldsymbol{\alpha}'$ is rescaled by $\rho^2_i$.

Finally, setting $\ell_i=\rho_i, \forall i=1,...,m$ implies the following interesting property.\\

\textbf{Corollary 1.2.}
\textit{Following from Theorem 1, and according to Corollary 1.1, if} $\boldsymbol{\ell}, \boldsymbol{\rho} \in \mathbb{R}_+^m : \ell_i=\rho_i=\sqrt{\xi_i}$,
\textit{then $k_{\mathcal{W}SE}(\boldsymbol{\alpha},\boldsymbol{\alpha}')$ turns into an exponential (aka Laplacian) kernel between the associated weight vectors, that is:
\begin{equation*}
    k_{\mathcal{W}SE}(\boldsymbol{\alpha},\boldsymbol{\alpha'})\big\vert_{\rho_i=\sqrt{\xi_i}}=k_{Exp}(\mathbf{a},\mathbf{a}')=e^{-\frac{\|\mathbf{a}-\mathbf{a}'\|_1}{\gamma}}
\end{equation*}
with $\gamma=4$.}\\

\textbf{Proof.} Consider the WSE kernel with $\mathcal{W}_2^2(\alpha,\alpha')$ computed with respect to the modified distance metric $\widetilde{d}^2(z_i,z_j')$. Since $\boldsymbol{\rho} \in \mathbb{R}_+^m$, we can write:
\begin{equation*}
    k_{\mathcal{W}SE}(\boldsymbol{\alpha},\boldsymbol{\alpha}')=e^{- \frac{\mathcal{W}_2^2(\boldsymbol{\alpha},\boldsymbol{\alpha}')}{2\boldsymbol{\rho}^2}} = e^{- \frac{1}{2}\sum_i\left(\sum_j \mathbf{C}_{ij}\mathbf{T}^*_{ij}\right)\frac{\xi_i}{\rho_i^2}}
\end{equation*}
Now, by setting $\rho_i=\sqrt{\xi_i}$ we obtain:
\begin{equation*}
    k_{\mathcal{W}SE}(\boldsymbol{\alpha},\boldsymbol{\alpha}')\big\vert_{\rho_i=\sqrt{\xi_i}}= e^{- \frac{1}{2}\sum_{ij} \mathbf{C}_{ij}\mathbf{T}^*_{ij}}
\end{equation*}
but, according to (\ref{eq:P1}), $\sum_{ij} \mathbf{C}_{ij}\mathbf{T}^*_{ij}$ is simply $\mathcal{W}_1(\boldsymbol{\alpha},\boldsymbol{\alpha}')$ computed under the binary ground metric, that is -- according to (\ref{eq:tv}) -- half of the Total Variation between $\boldsymbol{\alpha}$ and $\boldsymbol{\alpha'}$. Thus,
\begin{equation*}
    \begin{split}
    & k_{\mathcal{W}SE}(\boldsymbol{\alpha},\boldsymbol{\alpha}')\big\vert_{\rho_i=\sqrt{\xi_i}}= e^{- \frac{1}{2} \mathcal{W}_1(\boldsymbol{\alpha},\boldsymbol{\alpha'}) } = \\
    & = e^{- \frac{1}{2} \frac{TV(\boldsymbol{\alpha},\boldsymbol{\alpha'})}{2} } = 
    e^{- \frac{1}{2} \frac{\|\mathbf{a}-\mathbf{a}'\|_1}{2}} = \\
    & = k_{Exp}(\mathbf{a},\mathbf{a}') \text{ with } \gamma=4.
    \end{split}
\end{equation*}
\noindent
QED.\\

\textbf{Remark 1.} While Theorem 1 ensures that, under certain assumptions, it is possible to modify the ground metric to make the SE kernel on the weight vectors equal to a Wasserstein SE kernel -- under the modified ground distance -- the Corollary 1.2 reveals an unusual behaviour of $k_{\mathcal{W}SE}(\boldsymbol{\alpha},\boldsymbol{\alpha}')$: it can switch from $k_{SE}(\mathbf{a},\mathbf{a}')$ to $k_{Exp}(\mathbf{a},\mathbf{a}')$ for a given configuration of the vector-valued length-scale $\boldsymbol{\rho}$, that is:
\begin{equation*}
    \label{eq:kernel_switch}
    k_{\mathcal{W}SE}(\boldsymbol{\alpha},\boldsymbol{\alpha}') = 
    \begin{cases*}
        k_{Exp}(\mathbf{a},\mathbf{a}') \text{ with } \gamma=4 & if $\rho_i=\sqrt{\xi_i}$\\
        k_{SE}(\mathbf{a},\mathbf{a}') & otherwise
    \end{cases*}
\end{equation*}

This ``behavioural shift'' is particularly important, because SE and Exponential kernels are usually considered one the opposite of the other in terms of smoothness, with SE infinitely differentiable and Exponential not-differentiable.

\subsubsection{Equivalence via kernel's length-scale}
In the previous section we have proved that, in the case of univariate discrete probability measures, with the same support, and starting from a binary ground metric, it is possible to define a \textit{modified ground metric} leading to $\mathcal{W}_2^2(\boldsymbol{\alpha},\boldsymbol{\alpha}')=\|\mathbf{a}-\mathbf{a}'\|_2^2$.

Although \textit{ground metric learning} is a relevant research topic \citep{cuturi2014ground,heitz2021ground,huizing2022unsupervised}, the most common setting consists in fixing a ground metric suitable for the dataset and the task, and then using the Wasserstein distance.
Here we explain how it is possible to tune the length-scale $\boldsymbol{\ell} \in \mathbb{R}^m$ of the SE kernel to obtain $k_{SE}(\mathbf{a},\mathbf{a}')=k_{\mathcal{W}SE}(\boldsymbol{\alpha},\boldsymbol{\alpha}')$, without any modification of the ground metric.\\

\noindent
\textbf{Theorem 2.} \textit{Define the set $\mathcal{J}=\{i: a_i\neq a_i'\}$. Then, setting $\ell_i$ as follows:}
\begin{equation}
\label{eq:th2}
    \ell_i^2 = 
    \begin{cases*}
        *, >0 & if $i \notin \mathcal{J}$\\
        \frac{(a_i-a_i')^2}{\mathcal{W}_2^2(\boldsymbol{\alpha},\boldsymbol{\alpha}')} \rho^2 \vert \mathcal{J} \vert & otherwise
    \end{cases*}
\end{equation}
\textit{leads to $k_{SE}(\mathbf{a},\mathbf{a}')=k_{\mathcal{W}SE}(\boldsymbol{\alpha},\boldsymbol{\alpha}')$, with the SE and the WSE kernels being anisotropic and isotropic, respectively.}\\

\noindent
\textbf{Proof.}
\begin{equation*}
    \begin{split}
    & k_{SE}(\mathbf{a},\mathbf{a}')=e^{-\frac{1}{2}\sum_i\left(\frac{a_i-a_i'}{\ell_i}\right)^2}=\\
    & = e^{-\frac{1}{2}\sum_{i\notin \mathcal{J}}\left(\frac{a_i-a_i'}{\ell_i}\right)^2} e^{-\frac{1}{2}\sum_{i \in \mathcal{J}}\left(\frac{a_i-a_i'}{\ell_i}\right)^2} =\\
    & = e^{-\frac{1}{2}\sum_{i \in \mathcal{J}}\left(\frac{a_i-a_i'}{\ell_i}\right)^2}
    \end{split}
\end{equation*}

with $e^{-\frac{1}{2}\sum_{i\notin \mathcal{J}}\left(\frac{a_i-a_i'}{\ell_i}\right)^2}=1$ according to the definition of the set $\mathcal{J}$. Thus, setting $\ell_i^2$ as stated leads to:
\begin{equation*}
    \begin{split}
    & k_{SE}(\mathbf{a},\mathbf{a}')= e^{-\frac{1}{2}\sum_{i \in \mathcal{J}}\left[\frac{(a_i-a_i')^2\mathcal{W}_2^2(\boldsymbol{\alpha},\boldsymbol{\alpha}')}{(a_i-a_i')^2\rho^2 \vert\mathcal{J}\vert }\right]}=\\
    & = e^{-\frac{1}{2}\sum_{i \in \mathcal{J}}\left[\frac{\mathcal{W}_2^2(\boldsymbol{\alpha},\boldsymbol{\alpha}')}{\rho^2 \vert\mathcal{J}\vert}\right]} = e^{-\frac{1}{2} \vert \mathcal{J} \vert \left[\frac{\mathcal{W}_2^2(\boldsymbol{\alpha},\boldsymbol{\alpha}')}{\rho^2 \vert\mathcal{J}\vert}\right]} = \\
    & = e^{-\frac{\mathcal{W}_2^2(\boldsymbol{\alpha},\boldsymbol{\alpha}')}{2\rho^2}} = k_{\mathcal{W}SE}(\boldsymbol{\alpha},\boldsymbol{\alpha}').
    \end{split}
\end{equation*}

\noindent
QED.\\

Theorem 2 represents a crucial result of this paper. As previously mentioned in Sec. \ref{sec3.1}, the WSE kernel should be considered isotropic, so if we are able to identify a suitable value for $\rho$, then we can easily derive the vector-valued length-scale $\boldsymbol{\ell} \in \mathbb{R}_+^m$ for the equivalent anisotropic SE kernel, according to (\ref{eq:th2}). Moreover, the same equation also clearly states that every $\ell_i$ is a function of $\boldsymbol{\alpha}$ and $\boldsymbol{\alpha}'$, leading to the following important remark.\\

\textbf{Remark 2.} \textit{To every isotropic and stationary WSE kernel -- defined on univariate discrete probability measures having the same support -- it is associated an equivalent anisotropic and \textbf{non-stationary} SE kernel -- working on the associated weight vectors.}\\

On the other way round, one could be interested in setting the length-scale of an isotropic stationary SE kernel (i.e., $\ell \in \mathbb{R}_+$), and then obtain the equivalent WSE kernel. Replacing $\ell_i$ with $\ell$, in (\ref{eq:th2}), requires to also replace $\rho$ with $\rho_i$, due to the presence of the term $(a_i-a_i')$ in the equation. This leads to the following Theorem.\\

\textbf{Theorem 3.} \textit{If one of the kernels, SE or WSE, is chosen as isotropic and stationary, the equivalent counterpart, that is WSE or SE, respectively, will be anisotropic and non-stationary.} Proof directly stems from (\ref{eq:th2}) and previous considerations.\\

Finally, according to the consideration about the role of $\rho_i$ -- specifically equation (\ref{eq:W_rho}) -- we prove that the only important relationship for the equivalence between the two kernels is a non-stationarity relationship.\\

\textbf{Theorem 4.}  \textit{In the case of univariate discrete probability measures, with
the same support, for each anisotropic stationary SE kernel, defined on the weight vectors, exists an anistropic \textbf{non-stationary} WSE kernel, defined on the probability distributions, and viceversa. Equivalence is established by the following equation:}
\begin{equation}
    \label{eq:th4}
    \rho_i^2 = 
    \begin{cases*}
        *, >0 & if $i \notin \mathcal{I}$\\
        \frac{\sum_j \mathbf{C}_{ij}\mathbf{T}^*_{ij}}{\left(\frac{a_i-a_i'}{\ell_i}\right)^2 + \frac{Q}{\vert \mathcal{I}\vert}}  & otherwise
    \end{cases*}
\end{equation}
\textit{where the set $\mathcal{I}$ is defined as in (\ref{eq:setI}), that is $\mathcal{I}=\{i: \mathbf{C}_{ij}\mathbf{T}^*_{ij}\}$, and $Q$ is the quantity}:
\begin{equation}
    Q = \sum_{i\notin\mathcal{I}}\left(\frac{a_i-a_i'}{\ell_i}\right)^2
    \label{eq:Q}
\end{equation}\\

\textbf{Proof}. From equation (\ref{eq:W_rho}) and according to the definition of the set $\mathcal{I}$, we can write
\begin{equation*}
    \begin{split}
    \frac{\mathcal{W}_2^2(\boldsymbol{\alpha},\boldsymbol{\alpha}')}{\boldsymbol{\rho}^2} = & \sum_i\left(\sum_j \mathbf{C}_{ij}\mathbf{T}^*_{ij}\right)\frac{1}{\rho^2_i} = \\
    = & \sum_{i\in\mathcal{I}}\left(\sum_j \mathbf{C}_{ij}\mathbf{T}^*_{ij}\right)\frac{1}{\rho^2_i} + \underbrace{\sum_{i\notin\mathcal{I}}\left(\sum_j \mathbf{C}_{ij}\mathbf{T}^*_{ij}\right)\frac{1}{\rho^2_i}}_{=0} = \\ 
    = & \sum_{i\in\mathcal{I}}\left(\sum_j \mathbf{C}_{ij}\mathbf{T}^*_{ij}\right)\frac{\left(\frac{a_i-a_i'}{\ell_i}\right)^2 + \frac{Q}{\vert \mathcal{I}\vert}}{\sum_j \mathbf{C}_{ij}\mathbf{T}^*_{ij}} = \\
    = & \sum_{i\in\mathcal{I}}\left[\left(\frac{a_i-a_i'}{\ell_i}\right)^2 + \frac{Q}{\vert\mathcal{I}\vert} \right] = \sum_{i\in\mathcal{I}}\left(\frac{a_i-a_i'}{\ell_i}\right)^2 + \sum_{i\in\mathcal{I}} \frac{Q}{\vert\mathcal{I}\vert} = \\ 
    = & \sum_{i\in\mathcal{I}}\left[\left(\frac{a_i-a_i'}{\ell_i}\right)^2 + \frac{Q}{\vert\mathcal{I}\vert} \right] = \sum_{i\in\mathcal{I}}\left(\frac{a_i-a_i'}{\ell_i}\right)^2 + Q = \\ 
    = & \sum_{i\in\mathcal{I}}\left(\frac{a_i-a_i'}{\ell_i}\right)^2 + \sum_{i\notin\mathcal{I}}\left(\frac{a_i-a_i'}{\ell_i}\right)^2 = \sum_{i=1}^m\left(\frac{a_i-a_i'}{\ell_i}\right)^2 = \\ 
    = & \frac{\|\mathbf{a}-\mathbf{a}'\|^2_2}{\boldsymbol{\ell}^2}
    \end{split}
\end{equation*}
and it follows $k_{\mathcal{W}SE}(\boldsymbol{\alpha},\boldsymbol{\alpha}') = k_{SE}(\mathbf{a},\mathbf{a}')$.\\

\noindent
QED.\\

It is important to remark that, according to (\ref{eq:th4}), the quantity $Q$, computed as in (\ref{eq:Q}), it is equally distributed over $\vert\mathcal{I}\vert$ different $\rho_i^2$, with $i\in\mathcal{I}$. This is the easiest choice, but it is not unique.

\section{Gaussian Process regression over univariate discrete probability measures}\label{sec4}

As previously mentioned in Sec. \ref{sec2.1}, learning a GP on a dataset $\mathcal{D}$ usually means tuning the kernel's hyerparameters via MLE maximization.

For the sake of explanation, we recall here the equations (\ref{eq:gp}) of GP's predictive mean and variance, with a slightly modified notation to deal with probability measures and the WSE kernel:
\begin{equation}
    \label{eq:wgp}
    \begin{split}
    \mu({\mathbf{\boldsymbol{\alpha}}}) & = \widetilde{\mathbf{k}}(\boldsymbol{\alpha},\mathbf{A})^\top \left[\widetilde{\mathbf{K}}+\lambda^2\mathbf{I}\right]^{-1}\mathbf{y}\\
    \sigma^2({\mathbf{\boldsymbol{\alpha}}}) & = \widetilde{k}(\boldsymbol{\alpha},\boldsymbol{\alpha}') - \widetilde{\mathbf{k}}(\boldsymbol{\alpha},\mathbf{A})^\top \left[\widetilde{\mathbf{K}}+\lambda^2\mathbf{I}\right]^{-1} \widetilde{\mathbf{k}}(\boldsymbol{\alpha},\mathbf{A})
    \end{split}
\end{equation}

where $\mathbf{A}$ is a dataset of univariate probability measures, and $\widetilde{k}(\boldsymbol{\alpha},\boldsymbol{\alpha}')$ is a valid kernel between them. Finally, $\widetilde{\mathbf{k}}(\boldsymbol{\alpha},\mathbf{A})$ is a (column) vector whose $i$-th component is $\widetilde{k}(\boldsymbol{\alpha},\boldsymbol{\alpha}_i)$ -- and with $\boldsymbol{\alpha}_i$ the $i$-th row of the dataset $\mathbf{A}$ -- and $\widetilde{\mathbf{K}}$ is a symmetric $N \times N$ matrix with entries $\widetilde{\mathbf{K}}_{ij}=\widetilde{k}(\boldsymbol{\alpha}_i,\boldsymbol{\alpha}_j)$.\\

A first option is to simply choose $\widetilde{k}(\boldsymbol{\alpha},\boldsymbol{\alpha}') = k_{\mathcal{W}SE}(\boldsymbol{\alpha},\boldsymbol{\alpha}')$ and tune $\boldsymbol{\rho}$ via MLE maximization, as usual. However, as we empirically demonstrate here, if $k_{\mathcal{W}SE}(\boldsymbol{\alpha},\boldsymbol{\alpha}')$ is anisotropic, this procedure quickly leads to ill-conditioning of the kernel matrix $\mathbf{K}$, even for small 
dataset's size $N$. Consider the following setting:
\begin{itemize}
    \item the function (\ref{eq:test1}) defined in Sec \ref{sec2.3}, under a noise-free setting (i.e., $\lambda=0$)
    \item SE and WSE kernels, stationary and anisotropic, with $\boldsymbol{\ell},\boldsymbol{\rho} \in [10^{-5},1.6]^2$.
    \item $\sigma_f^2=1$ for both the two kernels
    \item seven points randomly sampled over the 1-dimensional probability simplex (i.e., seven probability distributions, with $m=2$)
\end{itemize}

\begin{figure}[h]
    \centering
    \includegraphics[scale=0.4]{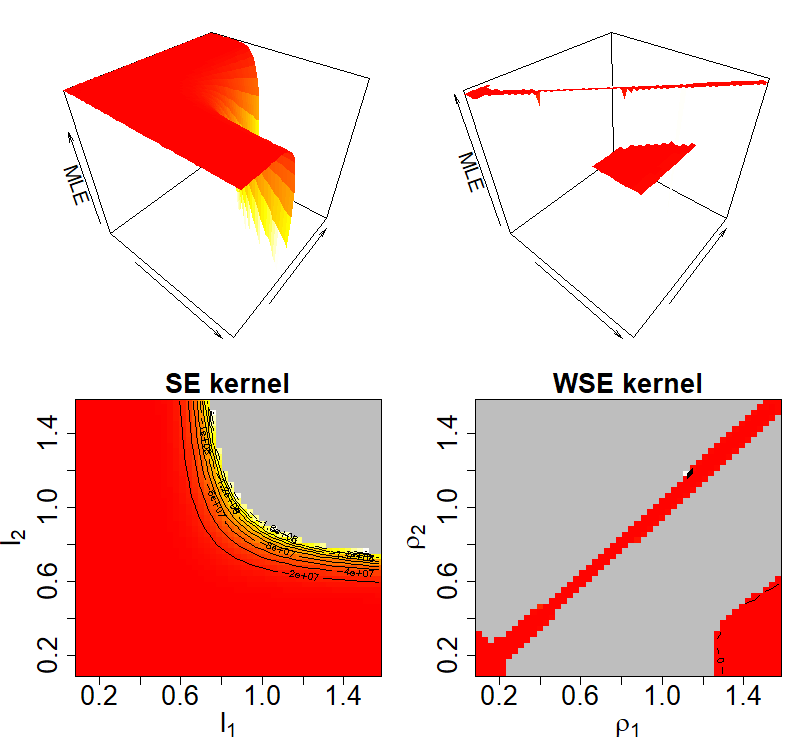}
    \caption{3D and contour plot of MLE for a SE (on the left) and a WSE (on the right) kernel. Both kernels are assumed stationary and anisotropic.}
    \label{fig:f6}
\end{figure}

Then, Figure \ref{fig:f6} shows the MLE with respect to the kernels' hyperparameters, respectively for SE kernel on the left-hand side and WSE kernel on the right-hand side. With just seven probability distributions (i.e., $N=7$), ill-conditioning occurs for both the kernels and MLE cannot be calculated for some $\boldsymbol{\ell}$ and $\boldsymbol{\rho}$ (i.e., grey areas). While MLE is defined over the most of the SE kernels' hyperparameters configurations, it basically does not exist for the WSE kernel.

\newpage

\begin{figure}[h]
    \centering
    \includegraphics[scale=0.4]{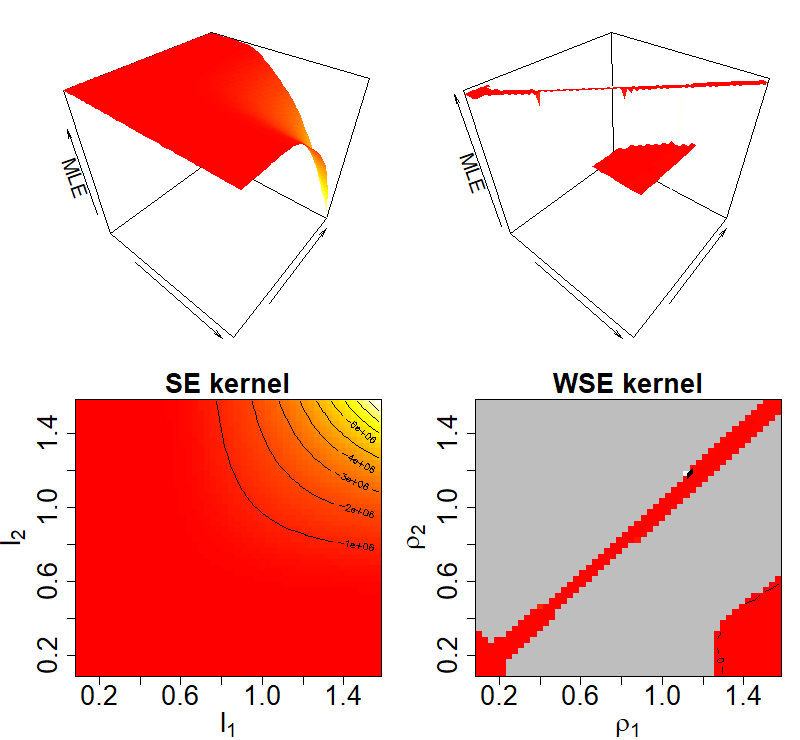}
    \caption{3D and contour plot of MLE, \textbf{after injecting additional noise} (i.e., nugget effect) for a SE (on the left) and a WSE (on the right) kernel. Both kernels are assumed stationary and anisotropic.}
    \label{fig:f7}
\end{figure}

When an artificial noise, $\lambda^2>0$, is added to overcome ill-conditioning, the expected result is that MLE can be calculated for all the considered configurations of $\boldsymbol{\ell}$ and $\boldsymbol{\rho}$. Thus we have decided to set $\lambda=10^{-5}$ but we have empirically observed that this workaround works only for the SE kernel, while it is definitely irrelevant for the WSE kernel, as depicted in Figure \ref{fig:f7}.

\newpage

Theorem 4 allows us to elegantly solve this numerial issue. Indeed, one can simply fit a GP with a SE kernel on the weight vectors of the univariate probability measures contained into the available dataset $\mathcal{D}$. Then, starting from the obtained length-scale $\boldsymbol{\ell}$, the Theorem 4 is used to obtain the associated $\boldsymbol{\rho}$ of the \textit{equivalent WSE kernel}.\\

Therefore, our approach consists into
choosing $\widetilde{k}(\boldsymbol{\alpha},\boldsymbol{\alpha}') = k_{\mathcal{W}SE}(\boldsymbol{\alpha},\boldsymbol{\alpha}')$ and computing $\boldsymbol{\rho}$ according to Theorem 4, making $k_{\mathcal{W}SE}(\boldsymbol{\alpha},\boldsymbol{\alpha}')$ anisotropic and \textbf{non-stationary} (we remark that, on the contrary, tuning $\boldsymbol{\rho}$ via MLE maximization assumes that the WSE kernel is stationary).\\

\begin{figure}[h]
\begin{subfigure}{.5\textwidth}
    \centering
    \includegraphics[scale=0.27]{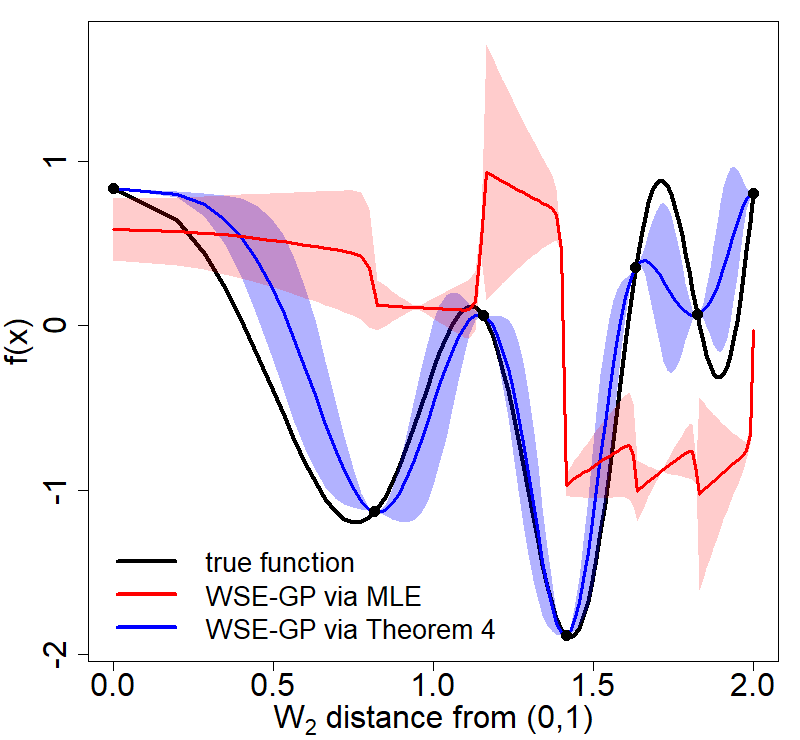}
\end{subfigure}
\begin{subfigure}{.5\textwidth}
    \centering
    \includegraphics[scale=0.27]{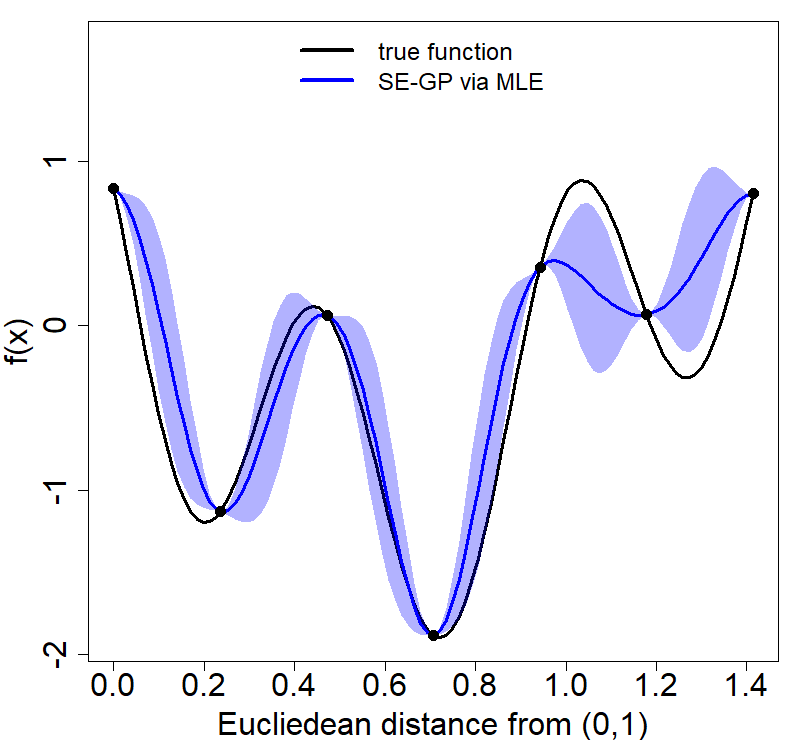}
\end{subfigure}
\caption{WSE lernel-based GP over the univariate probability simplex (on the left): differences between GP model learned via MLE maximization (in red) and via Theorem 4 (in blue). On the right, the SE kernel-based GP learned by assuming the probability simplex equipped with the Euclidean distance: Theorem 4 allows to map it to the equivalent WSE-GP model on the left.}
\label{fig:f8}
\end{figure}

Finally, using Theorem 4 or tuning $\boldsymbol{\rho}$ via MLE maximization (when possible) lead to two GP models with completely different shapes. Figure \ref{fig:f8} shows, on the left hand side chart, two GP models with WSE kernel learned through (blue) Theorem 4 (i.e., the \textit{equivalent WSE kernel}) and (red) MLE maximization (i.e., we have selected the $\boldsymbol{\rho}$ minimizing MLE, where calculable). In the chart, the probability simplex is equipped with the $\mathcal{W}_2$ distance. For completeness, on the right hand side chart, also the GP model with SE kernel is reported, over the probability simplex equipped with the Euclidean distance. the non-stationary mapping from SE to WSE kernel-based GP is evident.
\vskip -0.25cm

\newpage

\section{Experiments and results}\label{sec5}
In this section, we report results on a set of test problems, organized as follows:
\begin{itemize}
    \item four test problems on the \textbf{1-dimensional probability simplex}
    \begin{itemize}
        \item \textbf{test problem 02} (adapted from the \textit{Global Optimization Benchmarks and AMPGO} website\footnote{http://infinity77.net/global\_optimization/index.html}) -- it is a stationary function
        \item \textbf{test problem 13} (adapted from the \textit{Global Optimization Benchmarks and AMPGO} website) -- it is a convex, but non-stationary function
        \item \textbf{test problem 15} (adapted from the \textit{Global Optimization Benchmarks and AMPGO} website -- it is a smooth but non-stationary function
        \item \textbf{modified Xiong function} -- it is non-stationary test function specifically proposed in \citep{hebbal2021bayesian}.
    \end{itemize}
    \item one test problem on the \textbf{2-dimensional probability simplex}
    \begin{itemize}
        \item \textbf{Bird function} (adapted from the \textit{Global Optimization Benchmarks and AMPGO} website) -- it is a stationary function
    \end{itemize}
\end{itemize}

All the considered functions are global optimization test problems and have been adapted to be redefined over the probability simplex. All the details are reported in the Appendix \ref{secA1}.\\

To compare GP with SE and WSE kernel, we have set our experiments as follows, and separately for every test problem:
\begin{itemize}
    \item 20 data uniformly sampled at random over the probability simplex (i.e., 20 discrete probability measures with $m=2$ and $m=3$ for the 1-dimensional and the 2-dimensional probability simplex, respectively), and for which the function has been evaluated;
    \item 500 independent runs to mitigate the effect of randomness;
    \item 50 points uniformly sampled at random over the probability simplex and evaluated to calculate the root mean squared error (RMSE) between the predictions of the two GPs and the actual function.
\end{itemize}

To guarantee replicability of the experiments, both code (developed in R) and detailed results are shared at the following github repository:

\href{https://github.com/acandelieri/WaKer.git}{https://github.com/acandelieri/WaKer.git}\\

\noindent
As relevant results we report, for increasing dataset size (from $m+1$ to 20):
\begin{itemize}
    \item the percentage of runs in which learning the GP model failed due to ill-conditioning (i.e., MLE cannot be calculated);
    \item the RMSE between the true function and the GP's prediction, averaged over runs (if it was possible to learn a GP model on at least one run).
\end{itemize}

\begin{figure}[h]
\begin{subfigure}{.5\textwidth}
    \centering
    \includegraphics[scale=0.25]{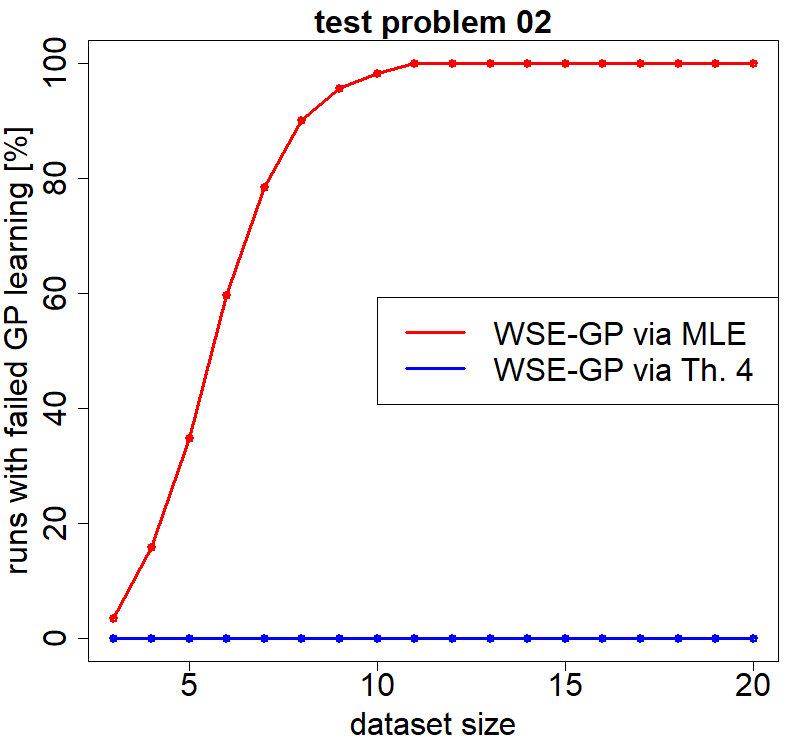}
\end{subfigure}
\begin{subfigure}{.5\textwidth}
    \centering
    \includegraphics[scale=0.25]{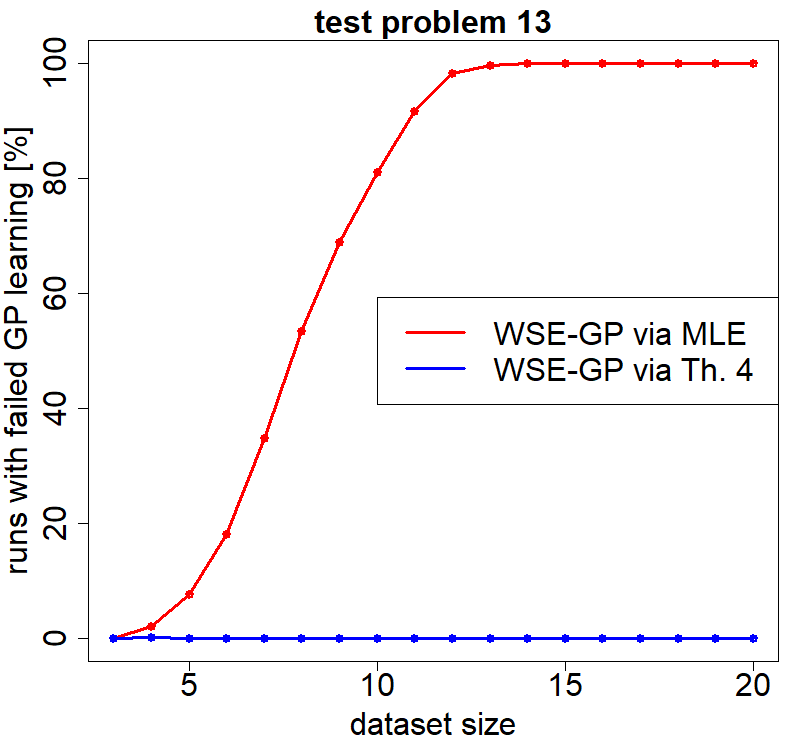}
\end{subfigure}
\begin{subfigure}{.5\textwidth}
    \centering
    \includegraphics[scale=0.25]{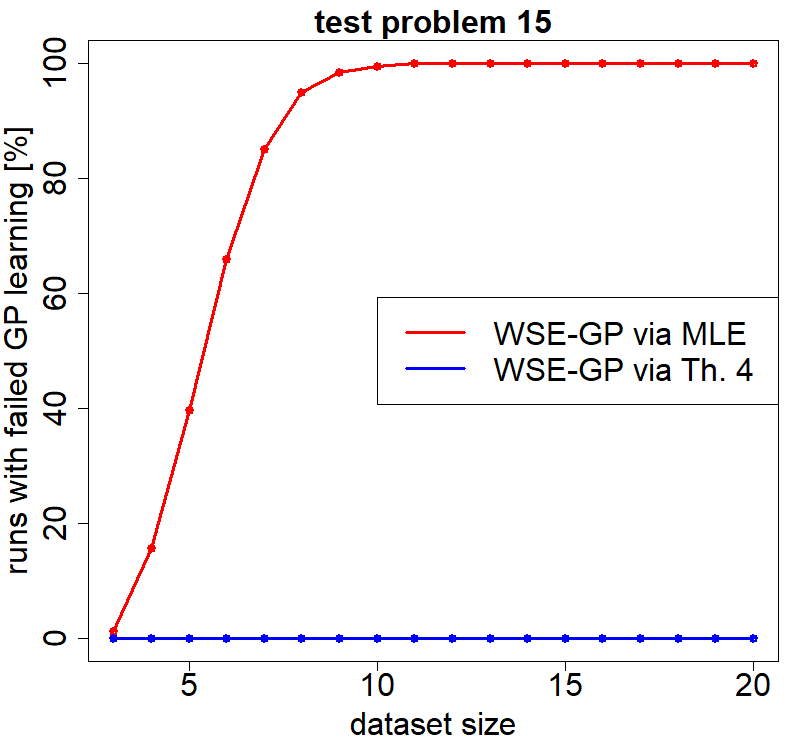}
\end{subfigure}
\begin{subfigure}{.5\textwidth}
    \centering
    \includegraphics[scale=0.25]{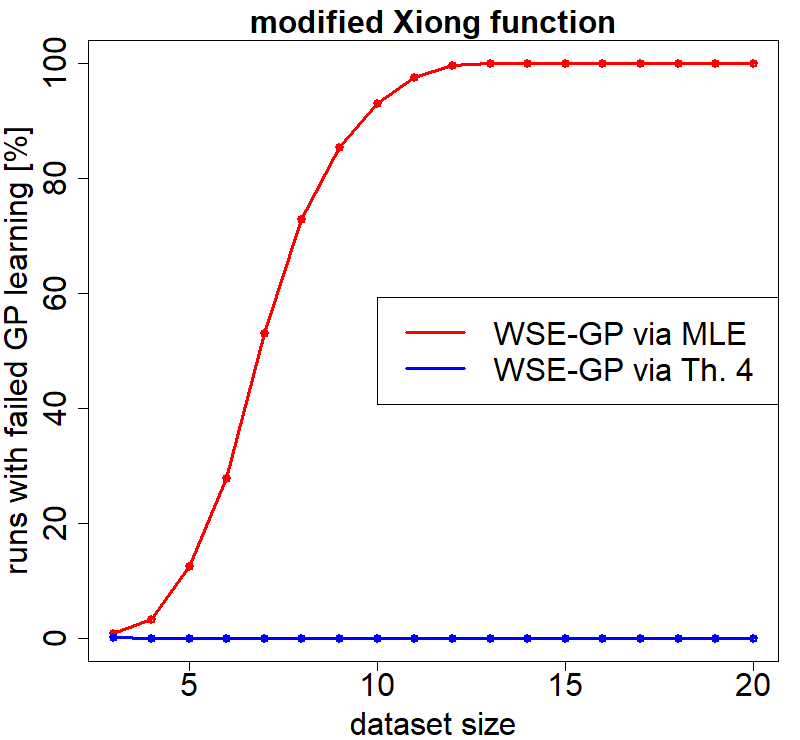}
\end{subfigure}
\caption{Percentage of independent runs in which GP learning was failed, separately for each 1-dimensional test problem. Comparison between WSE kernel based GP model learned via MLE maximization (red) and Theorem 4 (blue).}
\label{fig:f9}
\end{figure}

Figure \ref{fig:f9} shows the percentage of runs in which GP learning failed, for different dataset sizesa and separately for each test problem over the unidimensional probability simplex and for the two kernels. While adding additional noise (i.e., $\lambda^2>0$) allows to avoid ill-conditioning for the SE kernel, this does not work in the case of the WSE kernel, and the percentage of runs with failures quickly increases with the size of the dataset.

\newpage

\begin{figure}[h]
\centering
\includegraphics[scale=0.27]{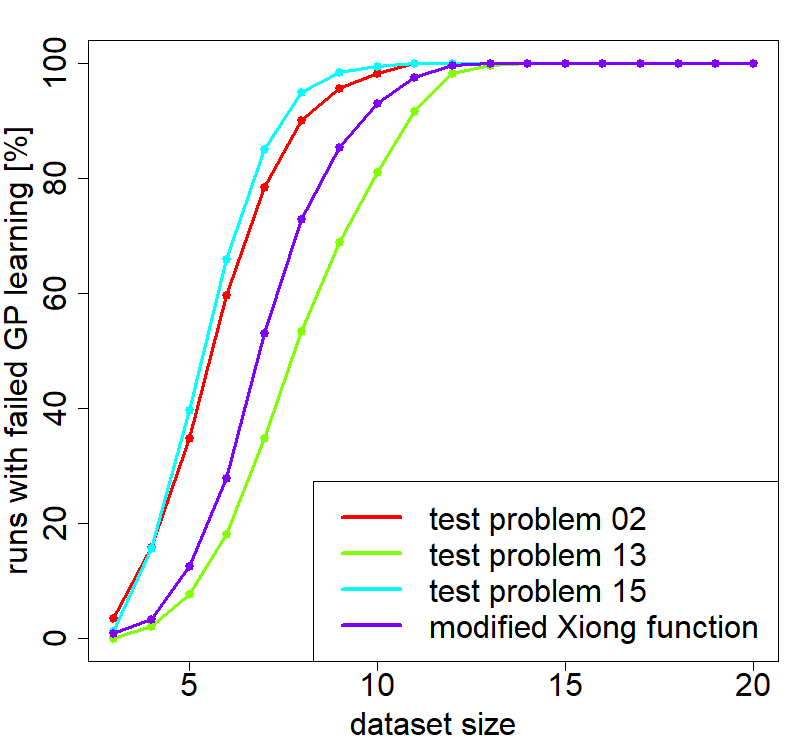}
\caption{Percentage of independent runs in which GP learning (i.e., MLE maximization) was failed: a comparison between 1-dimensional test problems.}
\label{fig:f10}
\end{figure}

Moreover, stationarity seems to be strictly linked to ill-conditioning  in the case of the WSE kernel. Figure \ref{fig:f10} clearly shows this relation, reporting the previous curves on a single chart: $100\%$ of runs with failures is achieved later for two out of the three non-stationary functions (i.e., test problem 13 and modified Xiong function).

\newpage

\begin{figure}[h]
\centering
\includegraphics[scale=0.27]{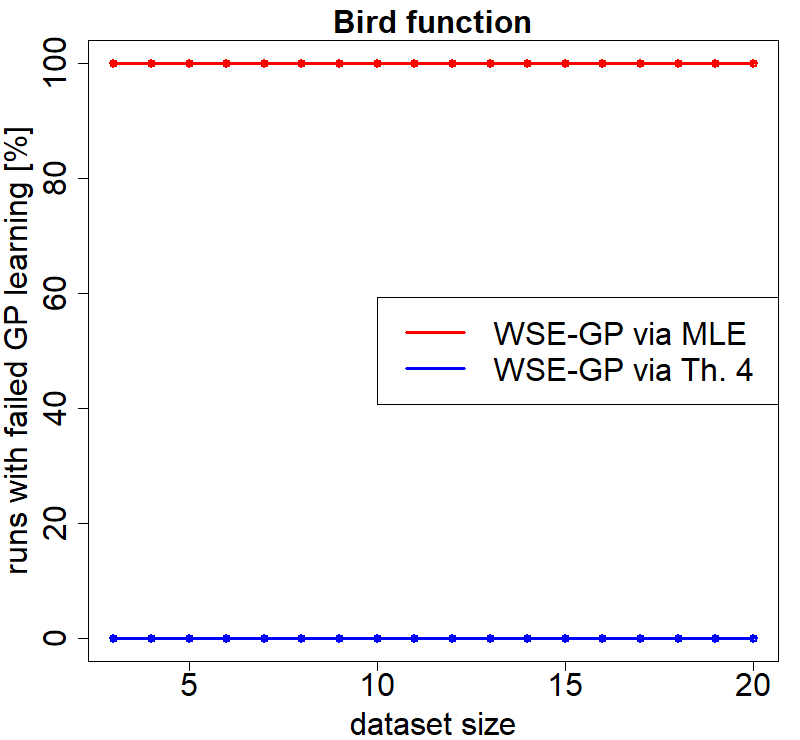}
\caption{Percentage of independent runs in which GP learning was failed on the 2-dimensional Bird function. Comparison between WSE kernel based GP model learned via MLE maximization (red) and Theorem 4 (blue).}
\label{fig:f11}
\end{figure}

As far as the experiment on the Bird function is concerned, the main result is that it was never possible to train a GP by maximizing the MLE of the WSE kernel, for any dataset size and over all the 500 runs (Figure \ref{fig:f11}).

\begin{figure}[h]
\begin{subfigure}{.5\textwidth}
    \centering
    \includegraphics[scale=0.24]{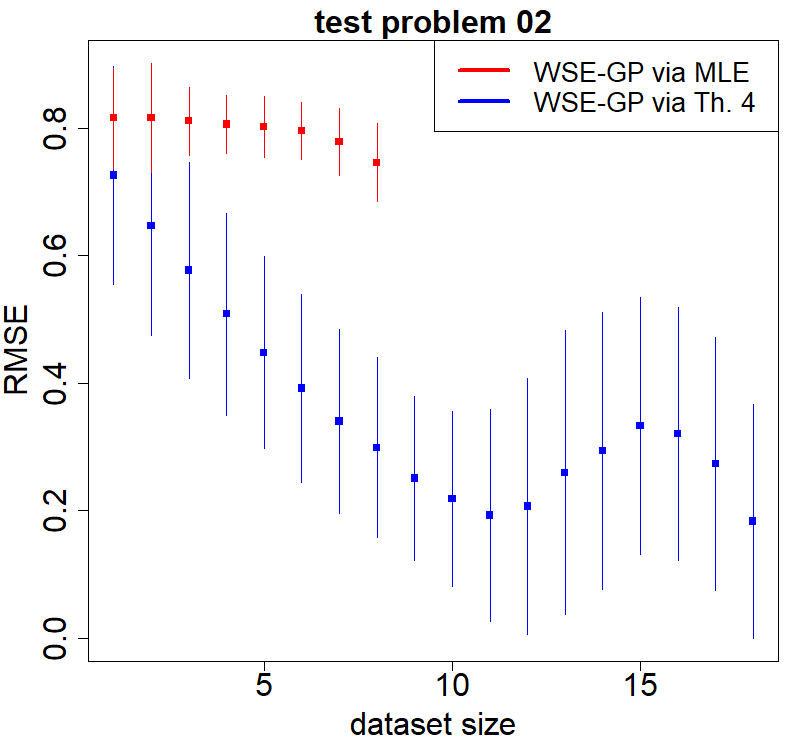}
\end{subfigure}
\begin{subfigure}{.5\textwidth}
    \centering
    \includegraphics[scale=0.24]{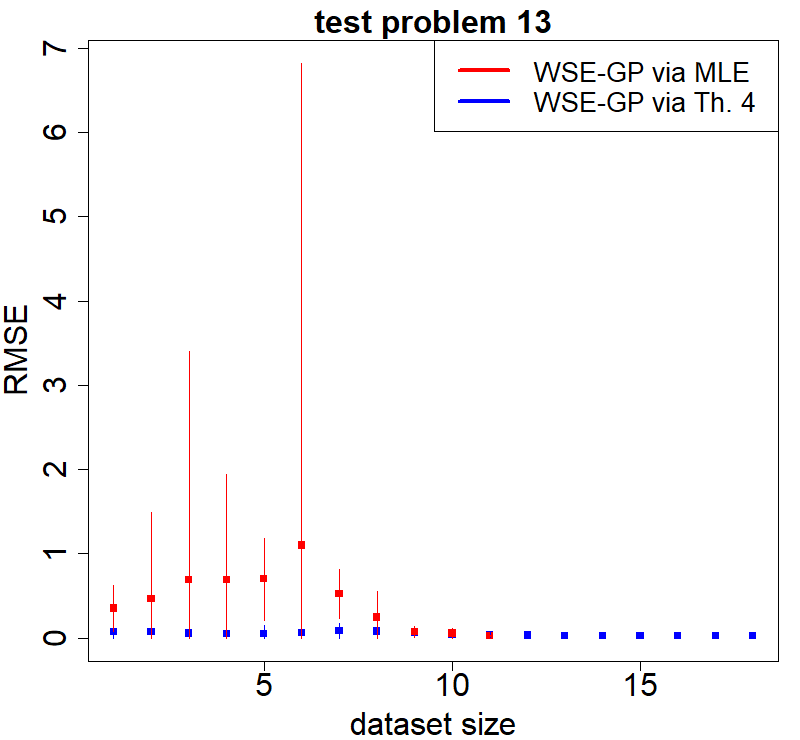}
\end{subfigure}
\begin{subfigure}{.5\textwidth}
    \centering
    \includegraphics[scale=0.24]{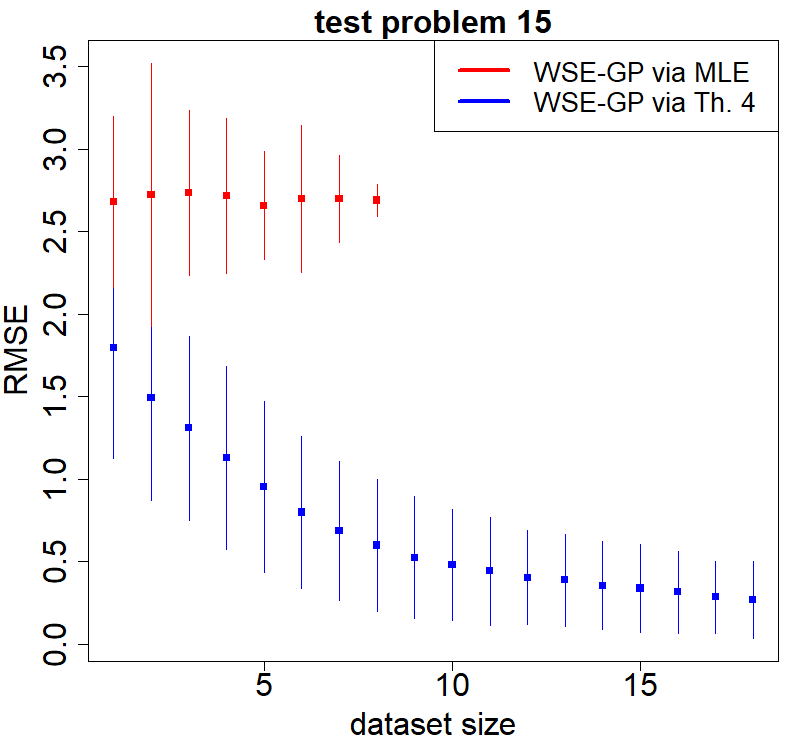}
\end{subfigure}
\begin{subfigure}{.5\textwidth}
    \centering
    \includegraphics[scale=0.24]{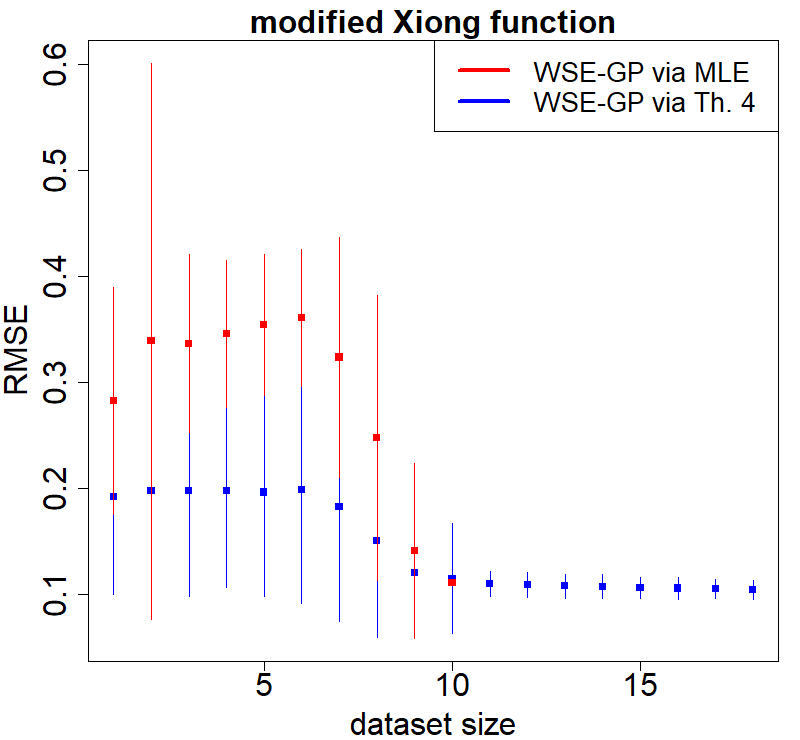}
\end{subfigure}
\caption{RMSE (average and standard deviation over 500 independent  runs) separately for each test problem: comparing WSE-GP learned via MLE maximization (red) and Theorem 4 (blue).}
\label{fig:f12}
\end{figure}

Finally, Figure \ref{fig:f12} shows the RMSE between the actual function and the GP's prediction with respect to the dataset size used to train -- if possible -- the GP model. Results refer only to the unidimensional test problems since the constant ill-conditioning for the WSE kernel on the Bird function.
These results prove that the GP learned via MLE maximization on the WSE kernel (red) is also less accurate than that obtained from the SE kernel-based GP and then mapped via Theorem 4. Finally, the difference between the RMSE of the two GPs is lower for two out of the three \textbf{non-stationary} test problems (i.e., test problem 13 and modified Xiong function).

\newpage

\section{Conclusions}\label{sec6}
We have addressed the problem of learning a Gaussian Process regression model on a dataset consisting of univariate discrete probability measures. The nature of the data requires to equip the input space with an appropriate distance. Specifically, the Wasserstein distance, $\mathcal{W}_2$, was considered in this paper. In the case of univariate probability measures $\mathcal{W}_2$ is Hilbertain and, consequently, guarantees that a Squared Exponential kernel based on it -- namely the Wasserstein Squared Exponential (WSE) kernel -- is Positive Definite.
However, numerical precision makes PD kernels behaving as PSD, leading to the well-known ill-conditioning issue affecting the GP learning procedure.

We have empirically demonstrated that this computational issue arises earlier and more frequently if the WSE kernel is used, especially if anisotropic. Furthermore, while adding a nugget effect allows to overcome ill-conditioning in the case of an Eucluidean SE kernel, it results completely irrelevant for the WSE kernel. As a consequence, the MLE maximization, at the core of the GP learning, cannot be used to fit a WSE kernel-based GP model.

The underlying motivation of this computational issue is in the non-Euclidean nature of the input space when equipped with the Wasserstein distance, as also reported in other research works. As relevant contribution, we have demonstrated that there exists a simple non-stationarity relation linking the WSE kernel and its Euclide SE counterpart, which can be exploited to elegantly solve the computational issue aforementioned. The GP model is learned assuming the input space as Euclidean and then the resulting model, along with its SE kernel, is mapped to the Wasserstein space through a simple algebraic transformation. This transformation does not exploit any structural properties of the input space (e.g., Riemannnian or pseudo-Riemannian structure exploited by log-exp maps); it is parametrized with respect to data and, therefore, it is in some sense learned form them. Computational advantages are evident from the reported experiments.\\

It is important to remark that the size of the dataset (i.e., the number of available univariate discrete probability measures, $N$) impacts (cubically) on the computational cost of the GP learning procedure, while the number of dimensions (i.e., the size of the support, $m$) has not any impact\footnotetext{We are aware that computing distances in high-dimensional spaces is an issue, but it is general and not GP regression specific.}. Therefore, in the case of univariate discrete probability measures with different supports one can always define a unique support $\mathbf{z}^*=\cup_{i=1}^N\mathbf{z}^{(i)}$ for all the probability measures of the dataset: this could lead to \textit{sparse} weight vectors (i.e., many components equal to 0), but it allows to use the proposed approach without any modification.\\

As far as limitations are concerned, our results are only valid for univariate probability measures. Indeed, for multi-variate probability measures the Wasserstein distance has not usually a closed form and it is not Hilbertian, so the WSE kernel is not PD. Dealing with multi-variate probability measures is significantly more complicated and an extension of our results is therefore not straightforward and requires further investigation.

Another perspective for possible future research activities is related to the extension of the analysis to also continuous probability measures, as well as other kernels, like those of the Matérn family.

Finally, we are planning to adopt the proposed approach into a GP-based BO framework. Although it is now straightforward to use our equivalent WSE kernel-based GP as a probabilistic surrogate model, the next step requires to optimize the \textit{acquisition function} -- based on the GP -- over a space equipped with the Wasserstein distance. Thus, traditional gradient-based algorithms are not well-suited -- due to the non-linearity of the space -- and other approaches, like \textit{gradient flow} should be considered. A very recent and interesting approach has been reported in \citep{crovini2022batch}, which could be considered as a principled starting point.

\newpage 
\begin{appendices}

 \section{test problems}\label{secA1}
\begin{figure}[h]
\centering
\begin{subfigure}{.4\textwidth}
    \centering
    \includegraphics[scale=.2]{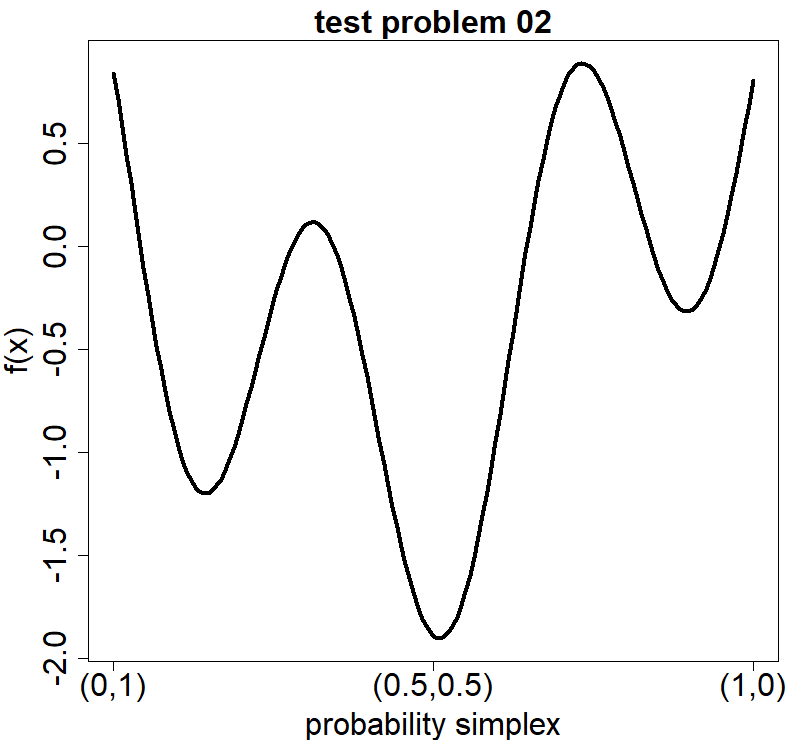}
\end{subfigure}
\begin{subfigure}{.4\textwidth}
    \centering
    \includegraphics[scale=.2]{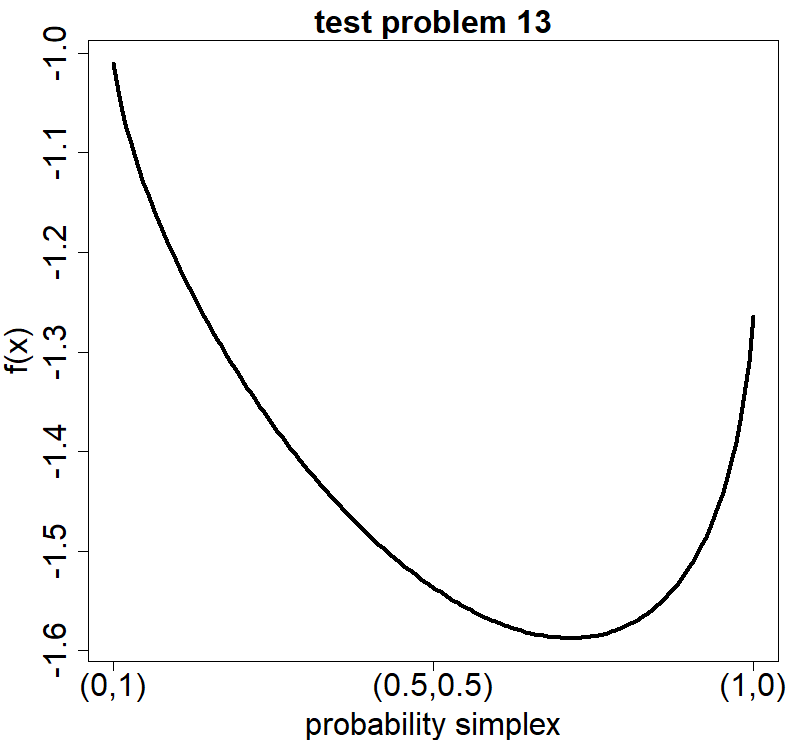}
\end{subfigure}
\begin{subfigure}{.4\textwidth}
    \centering
    \includegraphics[scale=.2]{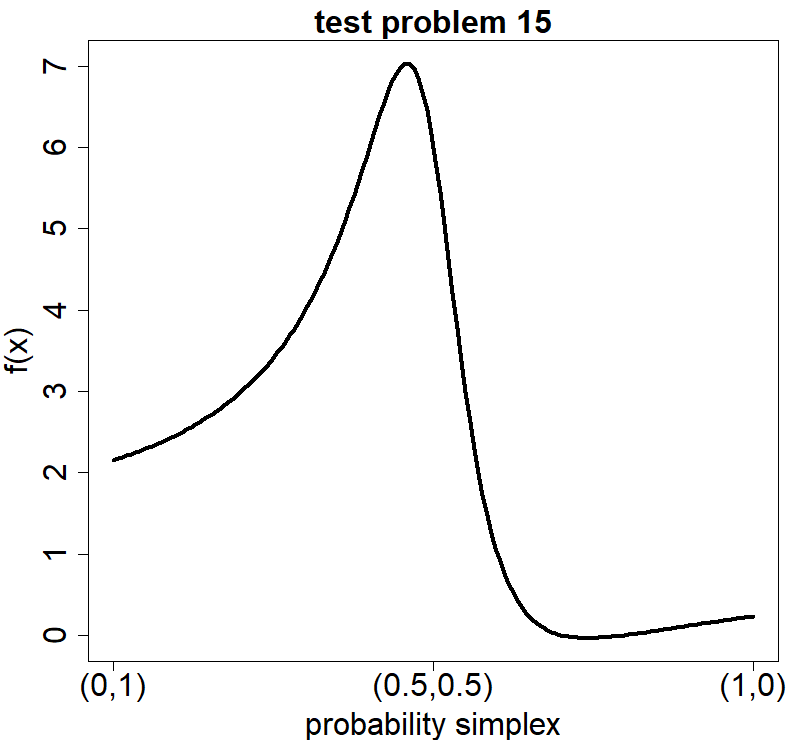}
\end{subfigure}
\begin{subfigure}{.4\textwidth}
    \centering
    \includegraphics[scale=.2]{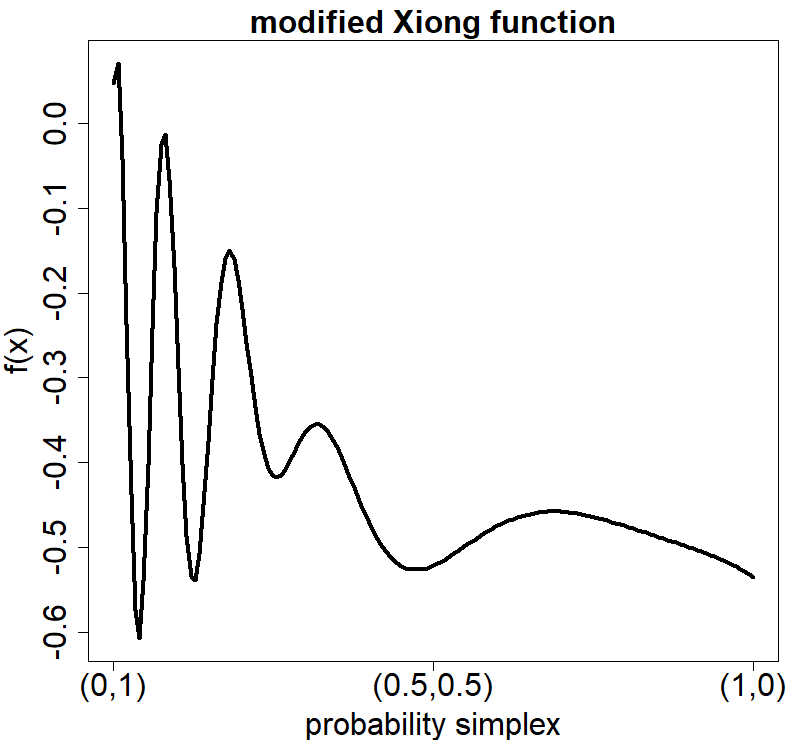}
\end{subfigure}
\caption{Graphical representation of the four 1-dimensional test problems.}
\label{fig:fA1}
\end{figure}

Figure \ref{fig:fA1} shows the four test problems defined over the univariate probability simplex. They are adaptation of well-known global optimization test functions. As follows, we report the adapted mathematical formulation:

\begin{itemize}
    \item \textbf{test problem 2} is the equation (\ref{eq:test1}) in the paper;
    \item \textbf{test problem 13} is defined as:
    \begin{equation*}
        f(x) = - \left(0.889x + 0.001\right)^\frac{2}{3} - \left[1-\left(0.889x + 0.001\right)^2\right]^\frac{1}{3}
    \end{equation*}
    \item \textbf{test problem 15} is defined as:
    \begin{equation*}
        f(x) = \frac{(10x - 5)^2 - 5 (10x-5) + 6}{ (10x-5)^2+1}
    \end{equation*}
    \item \textbf{modified Xiong function} is defined as:
    \begin{equation*}
        f(x) = -\frac{1}{2} \Bigg\{\sin\left[40(x-0.85)^4\right]\cos\left[2.5(x-0.95)\right] + \frac{x-0.9}{2} + 1\Bigg\} 
    \end{equation*}
\end{itemize}
For all the test problems, $x=a_1$ where $\mathbf{a}=(a_1,a_2)$.

\begin{figure}[h]
\centering
\includegraphics[scale=0.55]{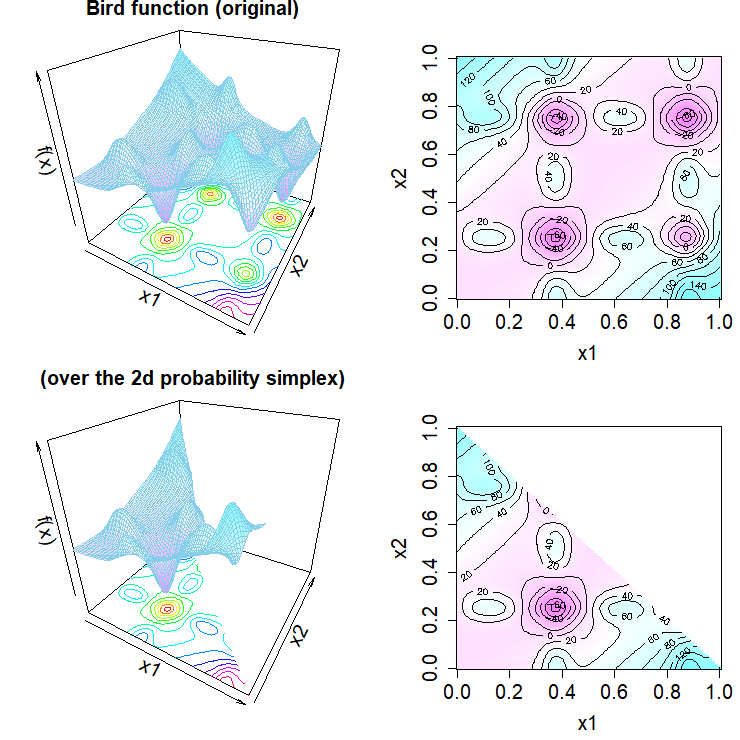}
\caption{Graphical representation of the Bird function: original test function (top) and restricted to the bi-dimensional probability simplex (bottom).}
\label{fig:fA2}
\end{figure}

The 2-dimensional problem is adapted from the Bird function. The original function is depicted on the top of Figure \ref{fig:fA2} (3D and contour plot), while the bottom of the figure shows the restriction to the 2-dimensional probability simplex.
The Bird function is defined as follows:
\footnotesize
\begin{equation*}
    f(\mathbf{x}) = [4\pi(x_1 - x_2)]^2 + [\cos (4\pi x_2 -2\pi)] e^{[1-\sin^2(4\pi x_1 - 2\pi)]} + [\sin (4\pi x_1 -2\pi)] e^{[1-\cos^2(4\pi x_2 - 2\pi)]} 
\end{equation*}
\normalsize

\newpage





\end{appendices}


\bibliography{sn-bibliography}


\end{document}